\documentclass[journal]{IEEEtran}

\usepackage{graphicx}
\usepackage{amssymb}
\usepackage{url}

\usepackage{url}
\usepackage{array}
\usepackage{times}
\usepackage{epsfig}
\usepackage{graphicx}
\usepackage{amsmath}
\usepackage{amssymb}
\usepackage{array}
\usepackage{multirow}
\usepackage{microtype}
\usepackage{algpseudocode}
\usepackage{bm}
\usepackage{framed}

\algdef{SE}[SUBALG]{Indent}{EndIndent}{}{\algorithmicend\ }%
\algtext*{Indent}
\algtext*{EndIndent} 
\usepackage{bbm}
\usepackage{float}

\begin{document}

\title{Partial Face Detection in the Mobile Domain}

\author{Upal~Mahbub,~\IEEEmembership{Student~Member,~IEEE,}
        Sayantan~Sarkar,~\IEEEmembership{Student~Member,~IEEE,}
        and~Rama~Chellappa,~\IEEEmembership{Fellow,~IEEE}
\thanks{The authors are with the Department of Electrical and Computer Engineering and the Center for Automation Research, UMIACS, University of Maryland, College Park, MD 20742.} 
\thanks{E-mail: \{umahbub, ssarkar2, rama\}@umiacs.umd.edu.}%
}

\maketitle

\begin{abstract}
Generic face detection algorithms do not perform well in the mobile domain due to significant presence of occluded and partially visible faces. One promising technique to handle the challenge of partial faces is to design face detectors based on facial segments. In this paper two different approaches of facial segment-based face detection are discussed, namely, proposal-based detection and detection by end-to-end regression. Methods that follow the first approach rely on generating face proposals that contain facial segment information. The three detectors following this approach, namely Facial Segment-based Face Detector (FSFD), SegFace and DeepSegFace, discussed in this paper, perform binary classification on each proposal based on features learned from facial segments. The process of proposal generation, however, needs to be handled separately, which can be very time consuming, and is not truly necessary given the nature of the active authentication problem. Hence a novel algorithm, Deep Regression-based User Image Detector (DRUID) is proposed, which shifts from the classification to the regression paradigm, thus obviating the need for proposal generation. DRUID has an unique network architecture with customized loss functions, is trained using a relatively small amount of data by utilizing a novel data augmentation scheme and is fast since it outputs the bounding boxes of a face and its segments in a single pass. Being robust to occlusion by design, the facial segment-based face detection methods, especially DRUID show superior performance over other state-of-the-art face detectors in terms of precision-recall and ROC curve on two mobile face datasets.
\end{abstract}

\begin{IEEEkeywords}
Facial Segment-Based Face Detection, Deep Convolutional Neural Network, Face Proposal Generation.
\end{IEEEkeywords}

\IEEEpeerreviewmaketitle

\section{Introduction}\label{Introduction}
\IEEEPARstart{R}{ecent} developments in convolutional neural network (CNN) architectures and the availability of large amount of face data have accelerated the development of efficient and robust face detection techniques \cite{RRanjan_Hyperface}\cite{CUHK_FD}\cite{YahooMultiview_FD}. State-of-the-art face detectors are mostly developed for detecting faces in unconstrained environments with large variations in pose and illumination \cite{Ramanan:2012:FDP:2354409.2355119}\cite{fddbTech}\cite{LFWTech}\cite{AFLWDataset}. However, development of face detectors optimized for detecting occluded and partially visible faces are becoming very essential because of the rapidly increasing usages of cameras mounted on mobile devices \cite{FSFD_Mahbub}\cite{Sarkar_DeepFeatureFD}. Reliable and fast detection of faces captured by the front camera of a mobile device is a fundamental step for applications such as active/continuous authentication of the user of a mobile device \cite{VMP_SPM_AA_2016}\cite{AA02_MahbubChellappa_BTAS2016}\cite{umd_Dataset}\cite{Mobio_2012}. 

Although, face-based authentication systems on mobile devices rely heavily on accurate detection of faces prior to verification, most state-of-the-art techniques are ineffective for mobile devices because of the following reasons:
\begin{enumerate}
\item The user's face captured by the front camera of the phone is, in many cases, only partially visible \cite{AA02_MahbubChellappa_BTAS2016}. While most modern face detectors such as Viola-Jones's \cite{VJFull}, DPM \cite{Ramanan:2012:FDP:2354409.2355119}, Hyperface \cite{RRanjan_Hyperface}, yahoo multiview \cite{YahooMultiview_FD}, CUHK \cite{CUHK_FD} etc. work well on detecting multiple frontal or profile faces at various resolutions, they frequently fail to single partially visible faces as they do not explicitly model partial faces.

\item For active authentication the recall rate needs to be high at very high precision. Many of the available face detectors have a low recall rate even though the precision is high. When operated at high recall, their precision drops rapidly because of excessive false positive detection  \cite{AA02_MahbubChellappa_BTAS2016}. 

\item  The algorithm needs to be simple, fast and customizable to operate in real-time on a cellular device. While in \cite{Sarkar_DeepFeatureFD} and \cite{AA_Samangouei_CNN} the authors deploy CNNs on mobile GPUs for face detection and verification, most CNN-based detectors, such as \cite{RRanjan_Hyperface} and some generic methods like \cite{Ramanan:2012:FDP:2354409.2355119} are too complex to run on the mobile platform.
\end{enumerate}

\begin{figure}[t]
\centering
\includegraphics[width=0.4\textwidth]{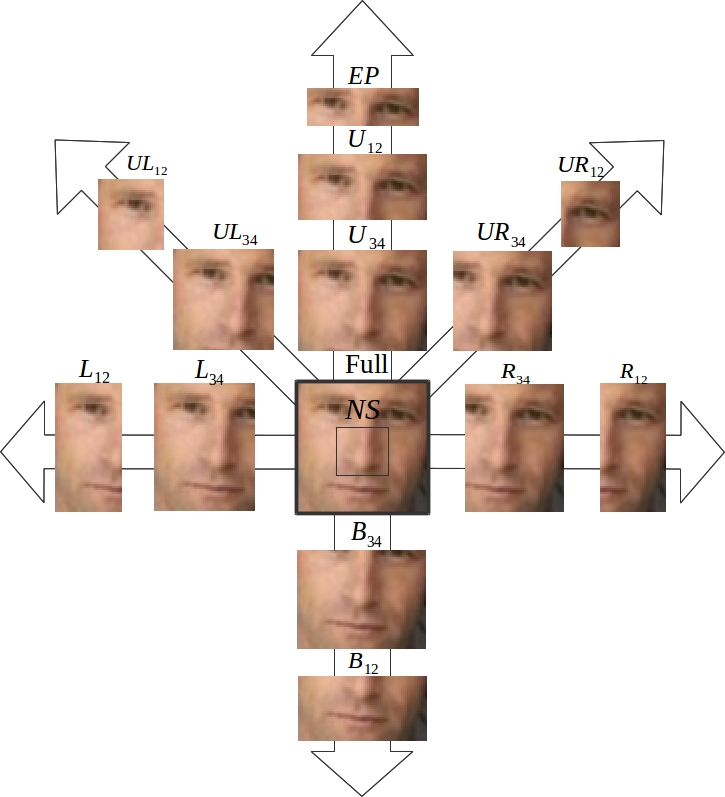}
\vskip -8pt
\caption{Decomposition of a full face into $14$ facial segments \cite{FSFD_Mahbub}.}
\label{FaceCrop}
\vskip -10pt
\end{figure}

On the other hand, images captured for active authentication offer certain advantages for the face detection problem because of its semi-constrained nature \cite{FSFD_Mahbub}. Usually there is a single user in the frame, hence there is no need to handle multiple face detection. The face is in close proximity of the camera and of high resolution, thus eliminating the need for detecting at multiple scales or resolution.

In this paper, we discuss the methods for facial segment-based face detection. Fig. \ref{FaceCrop} shows a sample decomposition of a full face into facial segments, detection of one or more of which might offer powerful clues about the whereabouts of the full face. Partial faces such as the ones present in images captured by the front camera of mobile devices can be handled if the algorithm is able to effectively combine detection of different facial segments into a full face. To address this requirement, three algorithms, namely, Facial Segment-based Face Detection (FSFD), SegFace and DeepSegFace, are presented in section  \ref{Proposal-based Methods}. These algorithms detect faces from proposals made of face segments by utilizing a fast proposal generation scheme that provides bounding boxes for faces and facial segments. FSFD and SegFace methods uses traditional feature extraction techniques and a support vector machine (SVM) classifier whereas DeepSegFace is a DCNN-based classifier for differentiating between proposals with and without faces.

While the three algorithms (FSFD, SegFace, DeepSegFace) presented here are fast and efficient, they have a lot of scope for improvement. A primary bottleneck of these algorithms is the proposal generation stage, which suffers from the following problems when trading off quality and speed:
\subsubsection{Slow Speed}The proposal generator can generate many proposals to ensure high recall, but it makes the pipeline slow as the detector must evaluate each of them. For example \cite{RRanjan_Hyperface} uses Selective Search \cite{SelectiveSearch}, which generates around $2000$ proposals per image, but the process is not real-time.
\subsubsection{Upper bound on Recall} If high recall rate is traded for speed, then one can use weak proposal generators which generate fewer proposals. However in this regime the detector is bound by the performance of the proposal generator and cannot detect faces in images where the proposal generator did not return any results. FSFD, SegFace and DeepSegFace use fast proposal generators (around 16 proposals per image), but have no way of recovering from the failures of proposal.
\subsubsection{Special Training} One way to generate a small number of proposals, yet have high recall, is to train specific proposal generators that identify faces and facial segments. However most off-the-shelf proposal generators detect generic objects, and one would have to retrain them for detecting faces.

Given the limitations of proposal-based detection, in this paper a regression-based end-to-end trainable face detector for detecting a single user face is proposed that does not require any proposal generation at all. This method, named, Deep Regression-based User Image Detector (DRUID), is a deep CNN-based face detector that returns not only the face bounding box, but also the bounding boxes of all the facial segments that are present along with the confidence scores for each segment in a single forward pass. DRUID utilizes a principled data augmentation technique to train on a relatively small number of image and it's throughput is very fast given its architecture and independence from the proposal generation stage. Moreover, training of DRUID is not done on a mobile face dataset similar to proposal-based approaches, yet, it performs significantly better than other methods mostly due to its unique architecture and data augmentation scheme. The augmentation also made it robust against scaling of faces and assists in finding the bounding boxes for faces of different sizes through regression during training. 

To summarize, this paper makes the following contributions:
\begin{enumerate}
\item Facial segment-based face detection techniques are explained from scratch.
\item Principles of three proposal-based methods for partial face detection, namely, FSFD, SegFace and DeepSegFace, are elaborated.
\item DRUID, a fast, novel regression-based deep CNN architecture, that detects partially visible and occluded faces without requiring any proposal generation is introduced.
\item Principled schemes are developed for augmentation and regularization of the classifiers and regressors.
\end{enumerate}

In section \ref{RelatedWorks} a summary of works done on face detection in general and in the mobile face domain in particular is given. In sections \ref{Proposal-based Methods} and \ref{Detection by Regression}, the proposal-based and the proposed end-to-end regression-based face detection techniques are described in detail. All the analysis and experimental results for the proposed methods and comparisons with state-of-the-art methods are provided in section \ref{Experimental Results}. Finally, a brief summary of this work as well as future directions of research are included in section \ref{Conclusion}.

\section{Related Works}\label{RelatedWorks}

\subsection{Approaches for Detection}
The detection problem in computer vision consists of predicting a bounding box for objects of interest in an image. Broadly, there are two approaches, one based on classification, the other on regression. In classification based approaches, one generates proposals or considers bounding boxes of multiple sizes at each location exhaustively and then classifies each proposal by deciding if an object is present or not. Popular methods in this category include \cite{Girshick:2014:RFH:2679600.2679851} and \cite{renNIPS15fasterrcnn}. In regression-based approaches, the deep network predicts the bounding box's location and dimensions through regression, without having to go though a proposal generation stage. Thus regression-based techniques are faster in general. Prime examples include \cite{redmon2016you} and \cite{liu2016ssd}

\subsection{General Purpose Face Detection}
Face detection, being one of the earliest applications of computer vision dating back several decades \cite{Bledsoe1965}\cite{Kanade1973}, was not applicable in real-world settings until 2004 because of poor performance in unconstrained  conditions  \cite{SurveyOnFD_CVIU2015}. The first algorithm that made face detection feasible in real-world applications was Viola and Jones's seminal work on boosted cascaded classification-based face detection \cite{VJFull}, which is still used widely in digital cameras, smartphones and photo organization software. However, researchers found that the method works reasonably well only for near-frontal faces under normal illumination without occlusion \cite{FDSurvey_MSR} and proposed extensions of the boosted architecture for multi-view face detection, such as  \cite{RotationInvMultiview_Huang}\cite{Multiview_heyden}. Even these extensions had shortcomings - they were difficult to train, and did not perform well because of inaccuracies introduced by viewpoint estimation and quantization \cite{FDSurvey_MSR}. A more robust face detector was introduced in \cite{Ramanan:2012:FDP:2354409.2355119} that used facial components or parts to construct a deformable part model (DPM) of a face. Similar geometrical modeling approaches are found in \cite{Component_Adv_Bileschi}\cite{LAEOdataset}. As support vector machines (SVMs) became effective for classification and robust image features like SURF, local binary pattern (LFP) histogram of oriented gradient (HoG) and their variants were designed, researchers proposed different combinations of features with SVM for robust face detection \cite{SurveyOnFD_CVIU2015}. In recent years, the performance of the original DPM-based method has been greatly improved \cite{HeadHunterMathias2014Eccv}. In the same paper, the authors introduced Headhunter, a new face detector that uses Integral Channel Features (ICF) with boosting to achieve state-of-the-art performance in face detection in the wild. In \cite{NPDDetector_2015}, the authors proposed a fast face detector that uses the scale invariant and bounded Normalized Pixel Difference (NPD) features along with a single soft-cascade classifier to handle unconstrained face detection. The NPD method is claimed to achieve state-of-the-art performance on FDDB, GENKI, and CMU-MIT datasets.

The performance improvement observed after the introduction of Deep Convolutional Neural Networks (DCNN) can be attributed to the availability of large labeled datasets, faster GPUs, the hierarchical nature of the deep networks and regularization techniques such as dropout\cite{SurveyOnFD_CVIU2015}. In \cite{YahooMultiview_FD}, the authors introduced a multi-task deep CNN architecture for multiview face detection which achieved state-of-the-art result on the FDDB dataset. Among other recent works, HyperFace \cite{RRanjan_Hyperface} is a deep multi-task framework for face detection, landmark localization, pose estimation, and gender recognition. The method exploits the synergy among related tasks by fusing the intermediate layers of a deep CNN using a separate CNN and thereby boosting their individual performances. Another multi-task learning framework is proposed in \cite{AllInOne_RRanjan} for simultaneous face detection, face alignment, pose estimation, gender recognition, smile detection, age estimation and face recognition using a single deep CNN. Unlike HyperFace \cite{RRanjan_Hyperface}, the method in \cite{AllInOne_RRanjan} utilizes domain-based regularization by training on multiple datasets in addition to AFLW and employs a robust, domain specific initialization.

\subsection{Face Detection for Active Authentication}
Continuous authentication of mobile devices requires detection and verification of the user's face, even if it is partially visible, to operate reliably \cite{AA02_MahbubChellappa_BTAS2016}.  \cite{Sarkar_DeepFeatureFD} and \cite{yang2015faceness} are two known methods that explicitly address the partial face detection problem. Specially in \cite{yang2015faceness} the authors achieve state-of-the-art performance on FDDB, PASCAL and AFW datasets by generating responses from face parts using an an attribute-aware deep network and refining the face hypothesis for partial faces.

\subsection{Active Authentication Datasets}\label{Dataset}	
Given the sensitive nature of smartphone usage data, there has been a scarcity of large dataset of front camera images in natural settings. However, the following datasets have been published in recent years that provide a platform to evaluate partial face detection methods in real-life scenarios.

\subsubsection{Constrained mobile datasets}
The MOBIO dataset \cite{Mobio_2012} contains $61$ hours of audio-visual data from a phone and laptop of $150$ participants spread over several weeks. However, since users were required to position their head inside an elliptical region within the scene while capturing the data, the face images of this dataset are constrained and do not represent real-life acquisition scenarios.

\subsubsection{Private datasets}
Google's Project Abacus data set consisting of $27.62$ TB of smartphone sensor signals collected from approximately $1500$ users for six months and Project Move dataset consisting of $114$ GB of smartphone inertial signals collected from $80$ volunteers over two months are the largest known datasets in this field \cite{NataliaNeverova_Abacus}\cite{AA02_MahbubChellappa_BTAS2016}. However due to privacy issues, these datasets are not publicly available. 

\subsubsection{Publicly available unconstrained mobile datasets}
The AA-01 dataset \cite{umd_Dataset} is a challenging dataset for front-camera face detection task which contains front-facing camera face videos for $43$ male and $7$ female IPhone users under three different ambient lighting conditions. In each session, the users performed five different tasks. To evaluate the face detector, face bounding boxes were annotated in a total of $8036$ frames of the $50$ users. This dataset, denoted as AA-01-FD, contains $1607$ frames without faces and $6429$ frames with faces \cite{FSFD_Mahbub}, \cite{Sarkar_DeepFeatureFD}. The images in this are semi-constrained as the subjects perform a set task during the data collection period. However they are not required or encouraged to maintain a certain posture, hence the dataset is sufficiently challenging due to pose variations, occlusions and partial faces.

The UMDAA-02 contains usage data of more than $15$ smartphone sensors obtained in a natural settings for an average of $10$ days per user \cite{AA02_MahbubChellappa_BTAS2016}. The UMDAA-02 Face Detection Dataset (UMDAA-02-FD) contains a total of $33,209$ images with manually annotated face bounding box, from all sessions of the $44$ users ($34$ male, $10$ female) sampled at an interval of $7$ seconds. This dataset is truly unconstrained, and hence challenging, as data is collected during real-time phone usage over a period of ten days on average per user. The face images have wide variation of pose and illumination, and it is observed that while the faces are mostly large in size, often the face images are only partially visible.

\section{Proposal-based Detection}\label{Proposal-based Methods}
\begin{figure*}[t]
\centering
\includegraphics[width=0.9\textwidth]{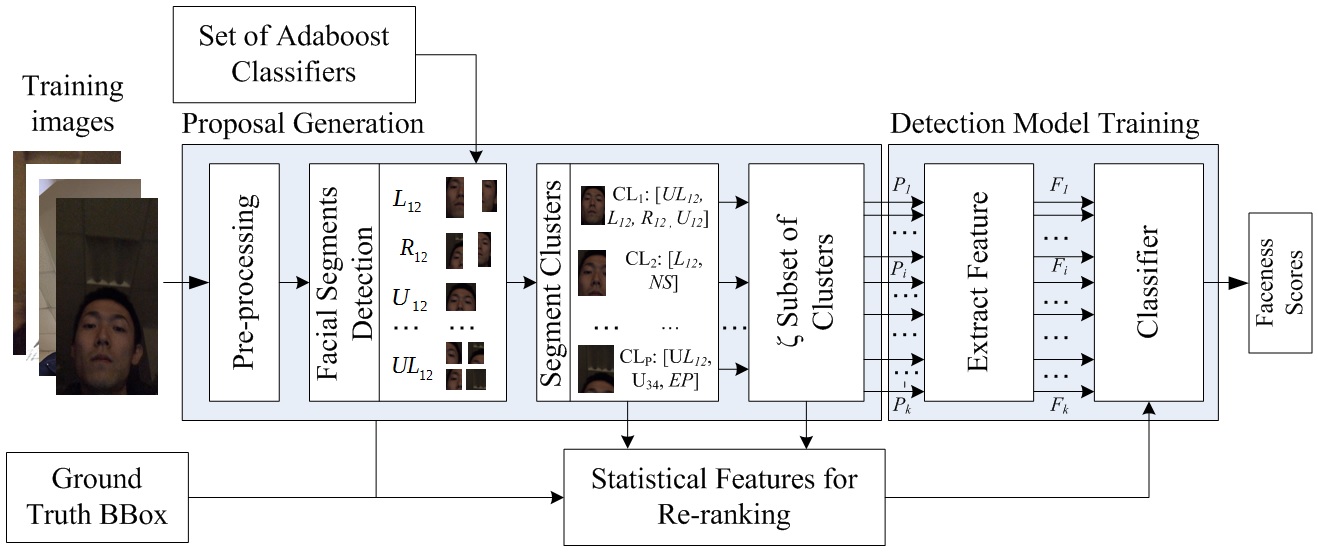}
\caption{Block diagram showing the general architecture of a face segment to face detector, with components such as facial segments-based proposal generator, feature extractor, classifier and re-ranking based on prior probabilities of segments \cite{SegFaceDeepSegFace_FG2017}.}
\label{SysteDiagramWithFeatures}
\vskip -8pt
\end{figure*}

In proposal-based detection methods, bounding boxes are generated as face proposals based on facial segments. The detectors are trained on the proposals and corresponding facial segment information to classify each segment as face/non-face with confidence scores. Three detectors of this type are presented along with a fast proposal generation mechanism. The detectors are 
\begin{enumerate}
\item Facial Segment-based Face Detection (FSFD) \cite{FSFD_Mahbub}
\item SegFace \cite{SegFaceDeepSegFace_FG2017}, and
\item DeepSegFace \cite{SegFaceDeepSegFace_FG2017}.
\end{enumerate}
Here, FSFD is a simple yet effective approach, SegFace has a more robust formulation and DeepSegFace is a deep CNN-based approach that is even more robust and accurate. A general pipeline for the proposal-based approach is shown in Fig. \ref{SysteDiagramWithFeatures}. As can be seen from the figure, the whole pipeline can be divided into two major parts - proposal generation and detection model training.

\subsection{Proposal generation}\label{section:PropGen}
For this work, $M=14$ facial segments are considered  which are Eye-pair ($EP$), Upper-left-half of face ($UL_{12}$), Upper-half of face ($U_{12}$), Upper-right-half of face ($UR_{12}$), Upper-left-three-fourth of face ($UL_{34}$), Upper-three-fourth of face ($U_{34}$), Upper-right-three-fourth of face ($UR_{34}$), Left-half of face ($L_{12}$), Left-three-fourth of face ($L_{34}$), Nose ($NS$), Right-three-fourth of face ($R_{34}$), Right-half of face ($R_{12}$), Bottom-three-fourth of face ($B_{34}$) and Bottom-half of face ($B_{12}$). An example of all the $14$ facial segments of a full face from the LFW dataset is shown in Fig. \ref{FaceCrop}. 

The set of facial segments is denoted by $S = \{FS_k \mid k=1,2,\dots M\}$, where $FS_k$ is the k-th facial segment. $M$ weak Adaboost facial segment detectors are trained to detect each of the segments in $S$. 

Given an image, all the segment detectors are employed. Each detector may return one or more facial segments for the same image. For each facial segment, the bounding box of the full face is estimated according to the ideal position of that segment relative to the whole face. For example, if the top left and bottom right corners of the bounding box obtained for segment $L_{12}$ are ($x_{1}^{L12}, y_{1}^{L12}$) and ($x_{2}^{L12}, y_{2}^{L12}$), respectively, then those for the estimated full face are ($x_{1}^{L12}, y_{1}^{L12}$) and ($min(w_{img}, x_{2}^{L12}+(x_{2}^{L12}-x_{1}^{L12})), y_{2}^{L12}$), where $w_{img}$ is the width of the image. The estimated face center from this segment is $(x_{2}^{L12}, y_{1}^{L12}+(y_{2}^{L12}-y_{1}^{L12})/2)$. For each estimated face center $j$, a cluster of segments $\text{CL}_j$ that depicts the full face is formed where, the other segments of that cluster have estimated face centers within a certain radius $r$ pixels from the center. Here, $j=\{1,2, \dots c_I\}$ and $c_I$ is the number of clusters formed for image $I$. 

A bounding box for the whole face $B_{\text{CL}_j}$ is calculated based on the constituent segments. For the generation of a proposal set, duplicate clusters that yield exactly same bounding boxes are eliminated and at most $\zeta$ face proposals are generated from each cluster by selecting random subsets of face segments constituting that cluster. Therefore, each proposal $P$ is composed of a set of face segments $S_P$, where $S_P \in \mathcal{P}(S)- \{\varnothing\}$ and $\mathcal{P}$ denotes the power set. To get better proposals, one can impose extra requirements such as $|S_P| > c$, where $|\cdot|$ denotes cardinality and $c$ is a threshold. Each proposal is also associated with a bounding box for the whole face, which is the smallest bounding box that encapsulates all the segments in that proposal.

Fig. \ref{SysteDiagramWithFeatures} depicts the integration of the proposal generation block into the face detection pipeline.

\subsection{Facial Segment-based Face Detection (FSFD)}\label{Subsec:FSFD}

\begin{figure*}[t]
\centering
\includegraphics[width=0.7\textwidth]{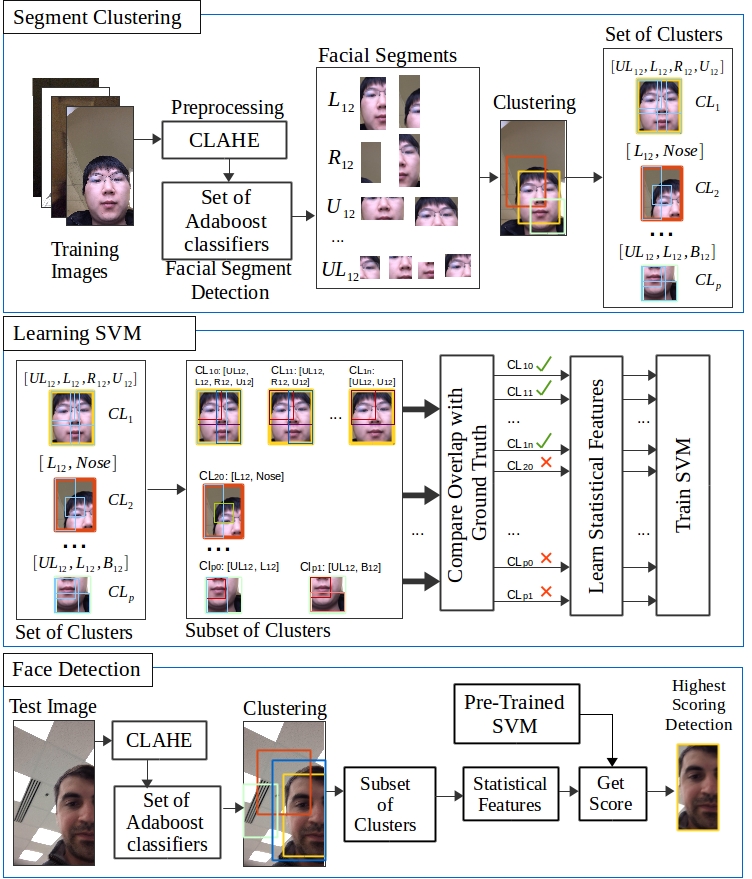}
\vskip -8pt
\caption{The block diagram of the overall system of FSFD \cite{FSFD_Mahbub}.}
\label{SysteDiagramSVM}
\vskip -10pt
\end{figure*}

The system block diagram for FSFD is shown in Fig. \ref{SysteDiagramSVM}. Segment clustering is done at first. Then, in the SVM learning phase, the first $\zeta$ subsets of the total set of facial segments from each cluster are regarded as proposed faces. Consider the $k$-th segment that was detected in an image. If there are $m$ segments that are within a certain distance from it and $m+1 \geq c$, then those $m+1$ segments are considered to be part of a single proposal. The value $c$ is the threshold that decides the minimum number of colocated segments that are required if they are to be declared to be a face proposal.

For example, if the $4$ segments $U_{34}$, $B_{12}$, $L_{12}$ and $UL_{12}$ form a cluster around $NS$ and $c=2$, then the viable subsets are $\allowbreak[[NS, U_{34}],\allowbreak [NS, B_{12}],\allowbreak \hdots,\allowbreak,\allowbreak [NS,\allowbreak U_{34}, \allowbreak B_{12},\allowbreak L_{12},\allowbreak UR_{34}]]$. The total number of subset here is $\sum_{j=1}^{4}{{4}\choose{j}}= 15$ including the complete set. Keeping the $k$-th segment fixed, $\zeta$ random subsets are considered for face proposals. In this example, $\zeta$ can vary from $1$ to $14$. Since for $m+1$ segments, the number of subsets is in the order of $2^{m+1}$, the number of subset is limited to $\zeta << 2^{m+1}$.

The bounding box of the face proposal is the smallest bounding box that contains all the estimated faces from all the facial segments in that proposal. Intuitively, larger the number of facial segments with better detection accuracy in a proposal, the higher the probability of that proposal being a face. Further, experimentally, it is found that some particular sets of facial segments are more likely to return a face than others, while some sets of segments provide more accurate bounding boxes with greater consistency than such other sets. 

A linear SVM classifier is trained on the proposed faces using the following prior probability values from the training proposal set that represents the likeliness of certain segments and certain combinations. These are

\begin{itemize}
\item Fraction of total true faces constituted by proposal $P$, i.e. 
\begin{equation}
\frac{| P \in \Theta^F |}{|\Theta^F |}, \nonumber
\end{equation} 
where $\Theta^F$ is the set of all proposals that return a true face.
\item The fraction of total mistakes constituted by proposal $P$, i.e. 
\begin{equation}
\frac{|P \in \Theta^{\overline{F}}|}{|\Theta^{\overline{F}}|}, \nonumber
\end{equation}, where, $\Theta^{\overline{F}}$ is the set of all proposals that are not faces.
\item For each of the $M$ facial segment $s_k \in S$, the fraction of total true face proposals of which $s_k$ is a part of, i.e.
\begin{equation}
\frac{|s_k \in S_p ; S_p \in \Theta^F|}{|\Theta^F|}, \text{ where, }k=1, 2, \hdots, M. \nonumber
\end{equation}
\item For each of the $M$ facial segment $s_k \in S$, the fraction of total false face proposals of which $s_k$ is a part of, i.e.
\begin{equation}
\frac{|s_k \in S_p ; S_p \in \Theta^{\overline{F}}|}{|\Theta^{\overline{F}}|},\text{ where, } k=1, 2, \hdots, M. \nonumber
\end{equation}
\end{itemize}

Experimentally, it is found that the nose detector is the most accurate of all the detectors, while $B_{12}$ is the least accurate. For $M$ facial segments, the size of this feature vector is $2M+2$ for each proposal. There are $2M$ values corresponding to the face and non-face probabilities of each of the $M$ segment and the remaining two values are the probabilities of the cluster being and not-being a face. Among the $2M$ values, only those corresponding to the segments present in the proposal are non-zero.

For each pre-processed test image, the proposed faces are obtained in a similar manner as the training faces. Thus, there are $\zeta$ face proposals from each face and feature vectors of size $2M+2$ for each proposal. The SVM classifier returns a confidence score for each proposal. The proposal that has the highest confidence score above a threshold is chosen as the detected face.

\subsection{SegFace}\label{Subsec:segFace}

SegFace is a fast and shallow face detector built from segments proposal. For each segment in $s_k \in S$, a classifier $C_{s_k}$ is trained to accept features $f(s_k)$ from the segment and generate a score denoting if a face is present. Output scores of $C_{s_k}$ are stored in an $M$ dimensional feature vector $F_{C}$, where, elements in $F_{C}$ corresponding to segments that are not present in a proposal are set to $0$. 

Another feature vector $F_S$ of size $2M+2$ is constructed using prior probability values as features as described in section \ref{Subsec:FSFD}. $F_C$ and $F_S$ are appended together to form the full feature vector $F$ of length $3M+2$. Then a master classifier $C$ is trained on the training set of such labeled vectors $\{F_i, Y_i\}$, where $Y_i$ denotes the label (face or no-face). Thus, $C$ learns how to assign relative importance to different segments and likeliness of certain combinations of segments occurring in deciding if a face is present in a proposal.
Thus, SegFace extends the face detection from segments concept in \cite{FSFD_Mahbub} using traditional methods to obtain reasonably accurate results. In our implementation of SegFace, HoG \cite{HOG_Features} features are used as $f$ and Support Vector Machine (SVM) classifiers \cite{SVM_tutorial} are used as both $C_{s_k}$ and $C$ for generating segment-wise scores, as well as the final detection score, respectively.

\subsection{DeepSegFace}
\begin{figure*}[t]
\centering
\includegraphics[width=0.9\textwidth]{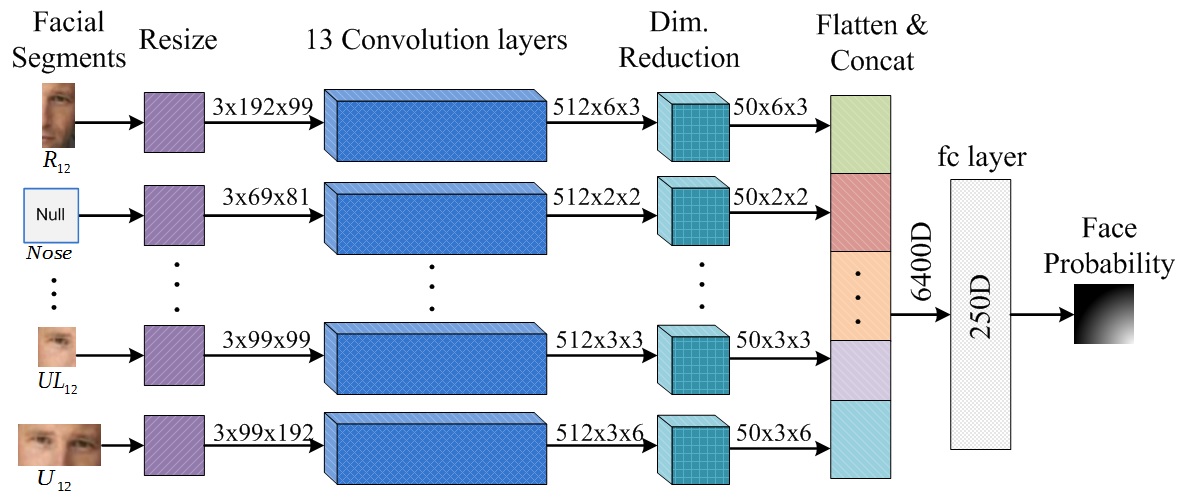}
\caption{Block diagram showing DeepSegFace architecture.}
\label{DeepSegFaceArch}
\end{figure*}

DeepSegFace is an architecture to integrate deep CNNs and segments-based face detection from proposals. At first, proposals, consisting of subsets of the $M=9$ parts as discussed earlier, are generated for each image. DeepSegFace is then trained to calculate the probability values of the proposal being a face. Finally, a re-ranking step adjusts the probability values from the network. The proposal with the maximum re-ranked score is deemed as the detection for that image.

The architecture of DeepSegFace is arranged according to the classic paradigm in pattern recognition: feature extraction, dimensionality reduction followed by a classifier. A simple block diagram of the architecture is shown in Fig. \ref{DeepSegFaceArch}. Different components of the figure are discussed here.

\emph{Convolutional Feature Extraction}: There are nine convolutional network columns, structurally similar to VGG16\cite{VGG} and initialized with its pretrained weights, for each of the nine segments. Thus each network has thirteen convolution layers arranged in five blocks. Each segment in the proposal is resized to standard dimensions for that segment, then the VGG mean value is subtracted from each channel. For segments not present in the proposal, zero-input is fed into the networks corresponding to those segments, as shown for the $Nose$ segment in the figure.

\emph{Dimensionality reduction}: The last convolutional feature map has $512$ channels, hence naively concatenating them results in a very large feature vector. Hence, a randomly initialized convolutional layer with filter size $1\times1$ and $50$ feature maps is appended to provide a learnable dimension reduction.

\emph{Classifier}: The outputs from the dimensionality reduction block for each segment-network are vectorized and concatenated to yield a $6400$-dimensional feature vector, constituents of which can be seen in Table \ref{DeepSegFaceConvLayers}. A fully connected layer of $250$ nodes, followed by a softmax layer of two nodes (both randomly initialized) is added on top of the feature vector. The two outputs of the softmax layer sum to one and correspond to the probability of the presence or absence of a face.

\emph{Re-ranking}: The DeepSegFace network outputs the face detection probabilities for each proposal in an image, which can be used to rank the proposals and then declare the highest probability proposal as the face in that image. However there is some prior knowledge that some segments are more effective at detecting the presence of faces than others. This information is available from the prior probability values discussed in section \ref{Subsec:segFace}. For DeepSegFace, these values are used to re-rank the final score by multiplying it with the mean of the statistical features.

\begin{table}
\centering
\caption{Structure of DeepSegFace's Convolutional layers (feature extraction and dimensionality reduction)}
\begin{tabular}{c  c c c c}
\hline
Segment         & Input &  Feature & Dim. Reduce & Flatten \\
\hline
\hline

$Nose$   & $3\times69\times81$ &  $512\times2\times2$  & $50\times2\times2$ & $200$ \\
\hline
$Eye$   & $3\times54\times162$ &  $512\times1\times5$  & $50\times1\times5$ & $250$\\
\hline
$UL_{34}$       & $3\times147\times147$ &  $512\times4\times4$  & $50\times4\times4$ & $800$\\
\hline
$UR_{34}$       & $3\times147\times 147$ &  $512\times4\times4$  &  $50\times4\times4$ & $800$\\
\hline
$U_{12}$       & $3\times99\times192$ &  $512\times3\times6$  &  $50\times3\times6$ & $900$\\
\hline
$L_{34}$       & $3\times192\times147$ &  $512\times6\times4$  & $50\times6\times4$ & $1200$\\
\hline
$UL_{12}$       & $3\times99\times99$ &  $512\times3\times3$  & $50\times3\times3$ & $450$\\
\hline
$R_{12}$       & $3\times192\times99$ &  $512\times6\times3$  & $50\times6\times3$ & $900$\\
\hline
$L_{12}$       & $3\times192\times99$ &  $512\times6\times3$  & $50\times6\times3$ & $900$\\

\hline

\hline
\end{tabular}
\label{DeepSegFaceConvLayers}
\vskip -5pt
\end{table}

\begin{table*}
\centering
\caption{Comparison of the proposed methods}
\begin{tabular}{p{2.5cm} p{4cm} p{4cm} p{4cm}}
\hline
Component    & FSFD     & SegFace &  DeepSegFace   \\
\hline
\hline
Proposal Generation 	& Clustering detections from cascade classifiers for facial segments & Clustering detections from cascade classifiers for facial segments & Clustering detections from cascade classifiers for facial segments \\
\hline
Low level features    & Prior probabilities & HoG and Priors &  Deep CNN features \\
\hline
Intermediate stage   & none    & SVM for segment $i$ outputs a score on HoG features of segment $i$ &  Dimension reduction and concatenation to single 6400D vector \\
\hline
Final classifier   & SVM trained on priors    & SVM trained on scores from part SVMs and priors &  Fully connected layer, followed by a softmax layer \\
\hline
Using priors   & Used as features in the final SVM    & Used as features in the final SVM &  Used for re-ranking of face probabilities in post processing\\
\hline
Trade offs    & Very fast but less accurate   & Fast but less accurate &  Slower but more accurate\\
\hline
\end{tabular}
\label{compareAlgos}
\vskip -5pt
\end{table*}

\emph{Facial Segment Drop-out for Regularization and Data Augmentation}: 	
	As mentioned in the proposal generation scheme, subsets of face segments in a cluster are used to generate new proposals. For example, if a cluster of face segments contains $n$ segments and each proposal must contain atleast $c$ segments, then it is possible to generate $\sum\limits_{k=t}^n {{n}\choose{k}}$ proposals.	
Now, if all the facial segments are present, the network's task is easier. However, all the nine parts are redundant for detecting a face, because of significant overlaps. Also often many segments are not detected by the weak segment detectors. Thus, one can interpret the missing segments as `dropped-out', i.e. some of the input signals are randomly missing (they are set to zero). Thus the network must be robust to face segments `dropping out' and generalize better to be able to identify faces.
Training with subsets of detected proposals also has the additional effect of augmenting the data. It has been observed that around sixteen proposals are generated per image. Many of these proposals are actually training the network to detect the same face using different combination of segments. 


\section{Detection by End to End Regression}\label{Detection by Regression}
For the regression method, a novel deep CNN-based method with a customized loss function, DRUID, will be discussed. DRUID generates facial segments as well as the full face bounding box with confidence scores and is completely end to end trainable, does not require proposal generation block and utilizes a unique data augmentation technique.

\subsection{DRUID: An End-To-End Network For Facial Segment-Based Face Detection}

\begin{figure}
\centering
\includegraphics[width=0.48\textwidth]{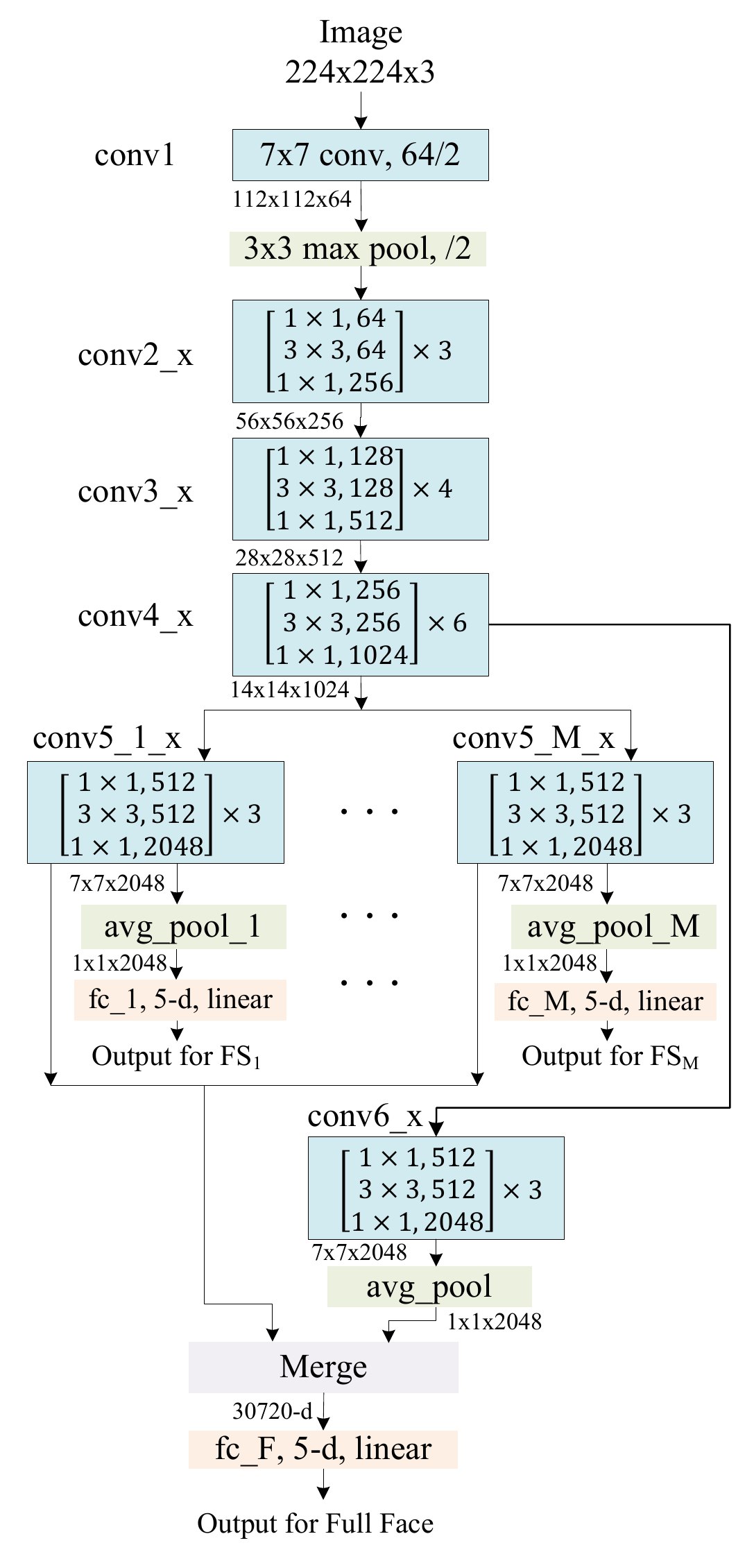}
\caption{Network Architecture for DRUID.}
\label{DeepSegE2ENetwork}
\end{figure}

DRUID is designed to detect a single face from an image along with all the $14$ facial segments (Fig. \ref{FaceCrop}) without any proposal generation. The algorithm takes a resized image as input and returns the location of each facial segment and the full face along with a visibility/confidence value. The network architecture of DRUID is shown in Fig.\ref{DeepSegE2ENetwork}. The network is made of Resnet units \cite{ResNet_CVPR}. Resnet like structures are built of basic blocks that  have two parallel paths, one that learns the residual and the other is either an identity transformation or a single convolution. For the residual path, a 'bottleneck' architecture is chosen, which consists of three sets of convolutional transformations (convolution, batch normalization, ReLU activation) of sizes $1\times1$, $3\times3$ and $1\times1$. These basic blocks are stacked into groups. Within a group, there is no downsampling. Let $convi\_x$ denote the $x^{th}$ block of the $i^{th}$ group. Each group starts with a block that downsamples by strided convolution (in both paths of the block), followed by other blocks with identity connections.

DRUID's architecture is shown in Fig. \ref{DeepSegE2ENetwork}. It has a common network trunk consisting of 3 Resnet groups, containing 3, 4 and 6 blocks respectively. Then the network is expanded to $M=14$ auxiliary branches, each containing the $conv5\_x$ unit of Resnet-50. Each of these auxiliary branches has $7\times7$ average pooling on top followed by a fully connected network ($fc\_1 - fc\_M$) with linear activation that gives a $10$-dimensional output. 

As for the full face detection output, an additional auxiliary branch is created without the fully connected layer (the convolution unit is denoted as $conv6\_x$ for this) and the flattened output of its average pooling block is merged with the intermediate flattened outputs of all the $conv5\_i\_x$ units (where $i=\{1, 2, \hdots, M\}$). The output of this merge block goes into another fully connected network $fc\_F$ with linear activation that outputs a $5$-dimensional vector.

\subsubsection{Loss Function for Training}

For the $i$-th facial segment, denote the estimated bounding box as $\mathbf{\widehat{b_i}}=\{\widehat{x_{1i}},\widehat{y_{1i}},\widehat{x_2i},\widehat{y_2i}\}$ and predicted visibility as $\widehat{v_i}$. Here, $\{\widehat{x_{1i}},\widehat{y_{1i}}\}$ denotes the top-left and $\{\widehat{x_{2i}},\widehat{y_{2i}}\}$ denotes the bottom right coordinates of the $i$-th facial segment.
 The ground truth bounding box for the $i$-th facial segment is $\mathbf{b_i}=\{x_{1i},y_{1i},x_{2i},y_{2i}\}$ and the ground-truth visibility is $v_i$. Also, the ground truth for the full face bounding box is $\mathbf{b^{F}}=\{ x_1^{F}, y_1^{F}, x_2^{F}, y_2^{F}\}$. Now, the $5$-dimensional vector that each of the fully connected layers predict by regression is $\{\widehat{x_{1i}},\widehat{y_{1i}},\widehat{x_2i},\widehat{y_2i}, \widehat{v_i}\}$.

The loss terms defined for this regression are
\begin{eqnarray}
L_{b} (i)&=& \parallel \mathbf{\widehat{b_i}}-\mathbf{b_i}\parallel _2^2 \\
L_{v} (i)&=& (\widehat{v_i}-v_i)^2 \\  
L_{x} (i)&=& (v_i \mathbbm{1}[\widehat{x_{1i}}-\widehat{x_{2i}}>0])^2 \\   
L_{y} (i)&=& (v_i \mathbbm{1}[\widehat{y_{1i}}-\widehat{y_{2i}}>0]) _2^2 \\    
L_{x_1}(i)&=& (v_i (x_1^{F}-\widehat{x_{1i}})\mathbbm{1}[x_1^{F}-\widehat{x_{1i}}>0])^2\\    
L_{x_2}(i)&=&(v_i (\widehat{x_{2i}}-x_2^{F})\mathbbm{1}[\widehat{x_{2i}}-x_2^{F}>0])^2\\    
L_{y_1} (i)&=& (v_i (y_1^{F}-\widehat{y_{1i}})\mathbbm{1}[y_1^{F}-\widehat{y_{1i}} >0])^2\\    
L_{y_2} (i)&=& (v_i (\widehat{y_{2i}}-y_2^{F})\mathbbm{1}[\widehat{y_{2i}}-y_2^{F}>0])^2\\    
L_{O} (i)&=& \parallel (1-IOU(\mathbf{\widehat{b_i}}, \mathbf{b_i})\mathbbm{1}[IOU(\mathbf{b_i},\mathbf{\widehat{b_i}}) \leq 0])\times \nonumber\\
&&\mathbbm{1}[v_i \neq 0])^2
\end{eqnarray}
where, $\mathbbm{1}$ denotes an indicator function. The total loss for each of the auxiliary network corresponding to facial segments and the loss for the full face detection network on top is defined as
\begin{eqnarray}
L_i=\sum_{k}\lambda_k L_k
\end{eqnarray}
where, $k\in \{v, x, y, x_1, x_2, y_1, y_2, O \}$ and $i \in S\cup\{F\}$.  Here, $F$ denotes the full face and corresponds to the topmost network. The components $L_b$ and $L_v$ are the $\ell_2$ losses for the predicted bounding box and visibility/confidence, respectively. Losses $L_x$ and $L_y$ imply that the coordinates of the bottom-right points are larger than those of the top-left points. $L_{x_1}$, $L_{x_2}$, $L_{y_1}$ and $L_{y_2}$ dictates that the coordinates of a facial segments are contained within the coordinates of the full face. Finally, $L_O$ implies that the overlap between the predicted bounding box and the ground truth bounding box for each segment should be as high as possible. Note that the ground truth visibility $v_i$ is a binary value for all $i\in S$ based on the visibility/non-visibility of corresponding segments in the image. However, $v_F$ is a floating point number which is calculated as $v_F=\frac{\sum_{i\in S}v_i}{|S|}$, i.e.  the fraction of total facial segments that is visible. Thus, a smaller value of $v_F$ will mean that only a few parts can be seen and indicate a partially visible face. $v_F$ is zero when no facial segments are visible, i.e. when there is no face. The $10$-dimensional output for each auxiliary branch indicates that each of the $i$ facial segments tries not only to predict the bounding box and visibility for that block itself, but also the bounding box and visibility of the full face in a manner that the corresponding loss is minimized.

\subsubsection{Data Augmentation}
\begin{figure}[t]
\centering
\includegraphics[width=0.48\textwidth]{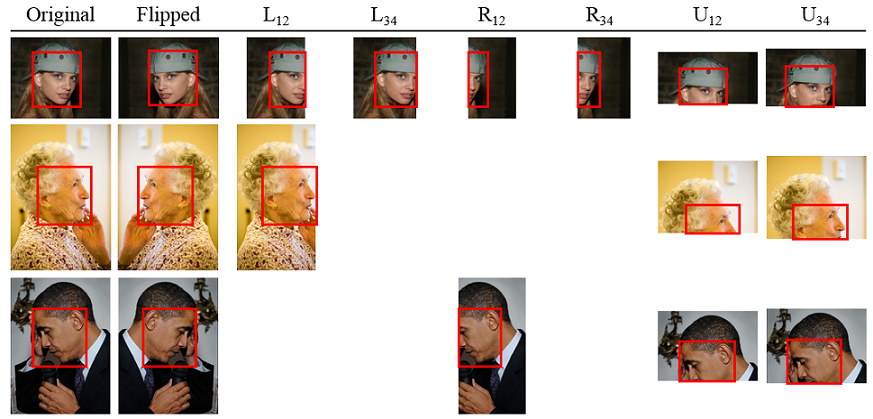}
\caption{Image processing for generating training data for DRUID ensuring partial visibility of faces.}
\label{DataAug}
\end{figure}

In order to ensure that the network is trained robustly for detecting facial segments, a novel data augmentation technique is adopted. Apart from flipping the input image and the corresponding bounding box coordinates, the images are cropped according to a pre-defined mechanism to ensure partial visibility in the training set. Some examples of such augmented data are shown in Fig. \ref{DataAug}. For the first row, since both the left, right and top portion are visible, there can be a maximum of six more augmented versions generated by cropping the image in a manner that only upto the $L_{12}$ or $L_34$ or $R_{12}$ or $R_{34}$ or $U_{12}$ or $U_{34}$ segment of the face is visible. Since, only left half and right half faces are visible in the original image in the second and third row of Fig. \ref{DataAug}, respectively, they can produce fewer number of augmented image.  The proposed data augmentation technique enables the network to train robustly for detecting partially visible faces with a few training images even if the original samples does not have many partially visible faces. In addition, the presence of certain segments dictate the presence of some other segments (e.g. if $L_{34}$ is visible then $L_{12}$ and $UL_{12}$ must be visible) and training with more partially visible faces enables the network to learn these interrelations as well.

Apart from augmentation, Gaussian blur is also applied on the original images with a probability of $0.7$ and a random radius between $0$ and $5$ pixels in order to improve the capability of the network to detect blurry faces. Moreover, a Gamma correction in the form of $I_o(x,y)=AI_i(x,y)^\gamma$ was applied to each pixel $I_i(x,y)$ of the normalized image, where, $A=1$ and $\gamma=2^s$ was set. For an image, the value of $s$ was obtained from a zero mean, unit variance Gaussian distribution. Thus, when $s$ is near the $0$, the transformed image is similar to the original, whereas, a positive value of $s$ would make the image darker and a negative value will make it brighter. Since there are quite a few very dark and very bright faces in the mobile face datasets, it is expected that this transformation would make DRUID robust against extreme illumination variations.

\subsubsection{Salient points of DRUID}
DRUID's architecture provides certain innate advantages which are discussed below:

\emph{Training Data}: DRUID is trained with images from AFLW, but for the task of mobile image detection. Therefore DRUID is able to handle the domain shift gracefully, without the need for specialized data from the mobile domain for training or fine-tuning as needed by methods like FSFD, SegFace and DeepSegFace.

\emph{Robustness}: DRUID is robust to occlusion due to its ability to detect partial faces which is ingrained in its architecture. DRUID is also very robust to variations in scale, if multi-scale data is present during training. DRUID does not need to scan the image at multiple scales to ensure it finds faces at all scales, since it does direct regression for the face location. Thus, as it has seen multi-scale faces during training, it is robust to scale variations to some degree. Also, because of the application of random Gamma transformation and Gaussian blur during training, DRUID performs well in detecting blurry and dark/bright faces, which are very frequent in mobile face domain, compared to other methods. It has the added advantage of multiple facial segment-based detection, which allows it to estimate the face bounding box with the help of one or more  segments even when all the other segment detectors fail. Hence, it is very robust against occlusion, illumination and pose variation.

\section{Experimental Results}\label{Experimental Results}
In this section, the experimental results of the facial segment-based detectors on the AA-01-FD and UMDAA-02-FD datasets are compared with a) Normalized Pixel Difference (NPD)-based detector \cite{NPDDetector_2015}, b) Hyperface detector \cite{RRanjan_Hyperface}, c) Deep Pyramid Deformable Part Model detector \cite{RRanjan_DeepPyramidFD}, and d) DPM baseline detector \cite{HeadHunterMathias2014Eccv}. 
\subsection{Experimental Setup}
FSFD, SegFace and DeepSegFace are trained on $3964$ images from AA-01-FD and trained models are validated using $1238$ images. The data augmentation process produces $57,756$ proposals from the training set, that is around $14.5$ proposals per image. The remaining $2835$ images of AA-01-FD are used for testing. For UMDAA-02-FD,  $32,642$ images are used for testing. In all experiments with FSFD, SegFace and DeepSegFace, $c=2$ and $\zeta=10$ are considered.

DRUID is trained using the Annotated Facial Landmarks in the Wild (AFLW)\cite{AFLWDataset} dataset which provides a large-scale collection of annotated face images gathered from Flickr, exhibiting a large variety in appearance (e.g., pose, expression, ethnicity, age, gender) as well as general imaging and environmental conditions. From a total of $21123$ faces that are manually annotated with up to $21$ landmarks per image, $18958$ images contain single faces, which are considered for training DRUID. The total number of training samples is $233029$ after data augmentation and negative sample extraction of which $100000$ are randomly chosen at the start of each epoch for training. Data is augmented by flipping ($18958$ images), cropping everything beyond $L_{12}$ ($15619$ images), $L_{34}$ ($13583$ images), $R_{12}$ ($16228$ images),  $R_{34}$ ($13493$ images), $U_{12}$ ($15362$ images) and $U_{34}$ ($15237$ images). $105591$ Negative samples are generated from the backgrounds of the training images with sizes equal to the size of the face in the corresponding image. 

The results are evaluated by comparing the ROC curve and precision-recall curves of these detectors since all of them return a confidence score for detection. The goal is to achieve high True Acceptance Rate (TAR) at a very low False Acceptance Rate (FAR) and also a high recall at a very high precision. Hence, numerically, the value of TAR at $1\%$ FPR and recall achieved by a detector at $99\%$ precision are the two metrics that are used to compare different methods.

\subsection{Selection of Parameter Values}
In our proposal generation scheme, $M=9$ is used. The nine parts under consideration are nose ($Nose$), eye-pair ($Eye$), upper-left three-fourth ($UL_{34}$), upper-right three-fourth ($UR_{34}$), upper-half ($U_{12}$), left three-fourth ($L_{34}$), upper-left-half ($UL_{12}$), right-half ($R_{12}$) and left-half ($L_{12}$. These nine parts, constituting the best combination $C_{best}$ \cite{FSFD_Mahbub} according to the analysis of effectiveness of each part in detecting faces, are considered in this experiment since the same adaboost classifiers are adapted in this work for proposal generation. The threshold $c$ is set to $2$. A small value is chosen to get high recall, at the cost of low precision. This lets one generate a large number of proposals, so that any face is not missed in this stage. $\zeta$ is set to $10$.

For DRUID, all the $14$ segments are considered. The parameters for training DRUID are set as follows: optimizer - adam, learning rate $0.0001$, $\beta_1=0.9$, $\beta_2=0.999$, $\epsilon=1e-08$, decay=$0.0$, total epoch $25$ and batch size $32$.

\begin{figure}[t]
\centering
\includegraphics[width=0.48\textwidth]{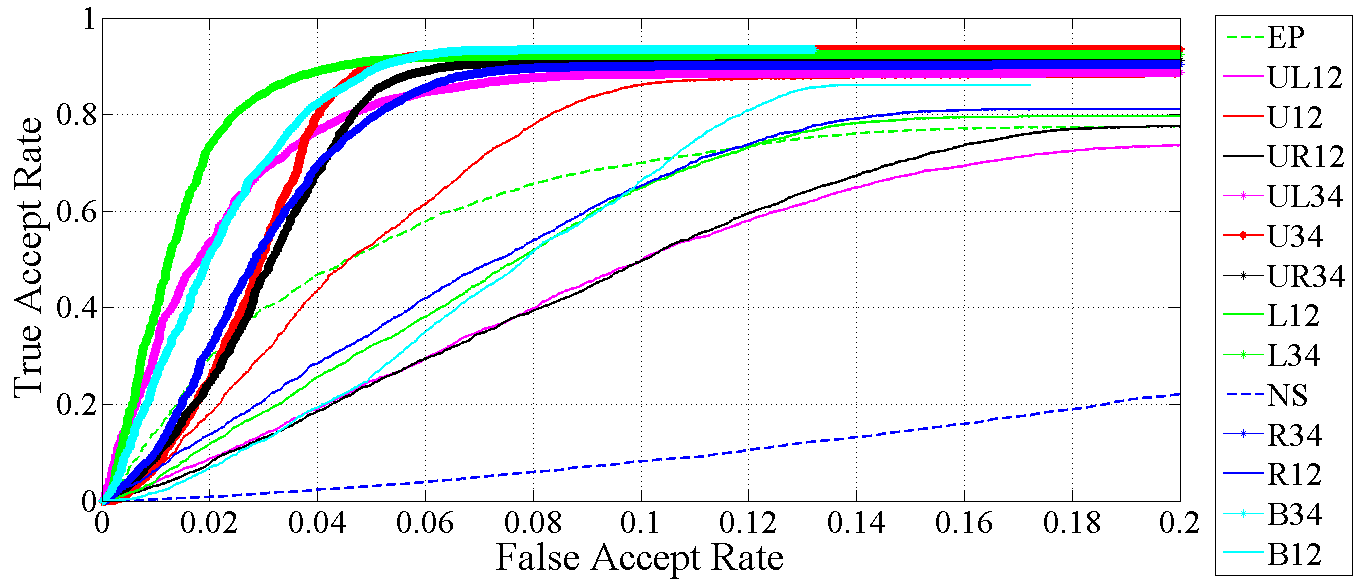}
\caption{TAR vs. FAR ROC plots for different segment prediction results obtained for DRUID on $18958$ images of the AFLW dataset.}
\label{TAR_FAR_ROCforAFLWSegments}
\end{figure}

\begin{figure}[t]
\centering
\includegraphics[width=0.48\textwidth]{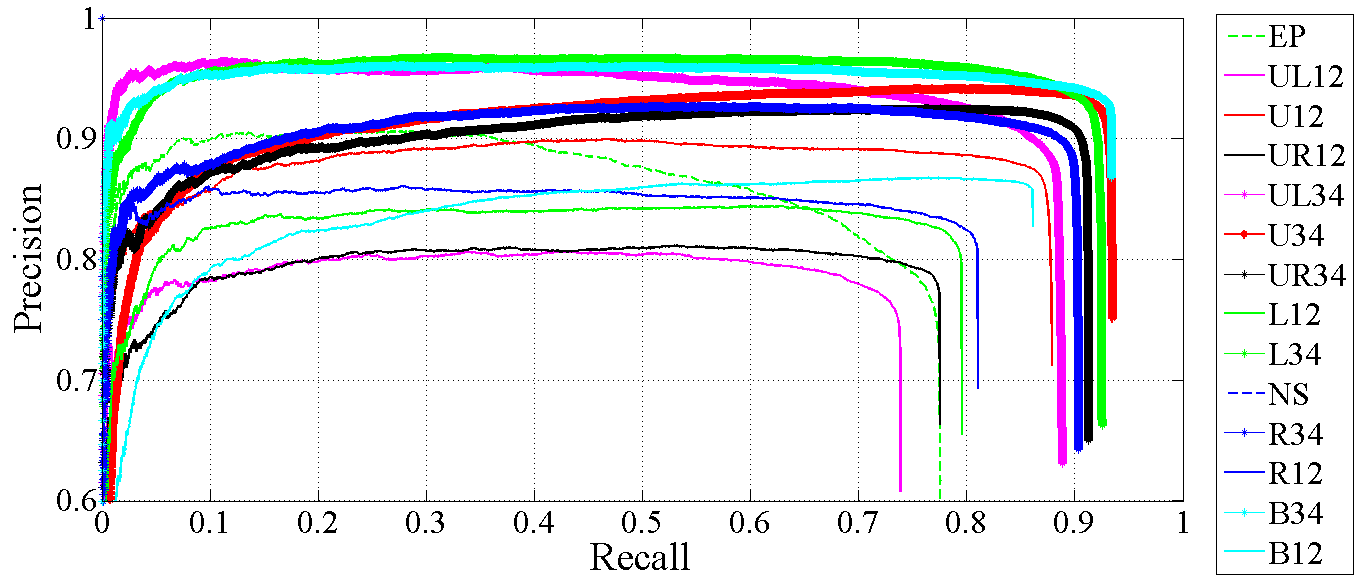}
\caption{Precision vs. Recall curves for different segment prediction results obtained for DRUID on $18958$ images of the AFLW dataset.}
\label{Precision_Recall_ROCforAFLWSegments}
\end{figure}

\subsection{Quantitative Analysis}

\begin{table}
\centering
\caption{Comparison at $50\%$ overlap on AA-01-FD and UMDAA-02-FD datasets}
\vskip -8pt
\begin{tabular}{c c c c c} 
\hline
\multirow{2}{*}{Methods} &
  \multicolumn{2}{c}{AA-01} &
  \multicolumn{2}{c}{UMDAA-02}\\
\cline{2-5}
	& TAR at 	& Recall at  & TAR at & 	Recall at	\\ 
	    & $1\%$ FAR	& $99\%$ Prec.	& $1\%$ FAR & $99\%$ Prec.	\\ 
\hline
\hline
NPD \cite{NPDDetector_2015}					& $29.51$ 	& $11.0$ 	& $33.49$ & $26.79$\\ 
\hline
DPMBaseline	\cite{HeadHunterMathias2014Eccv}			& $85.08$	& $83.25$ & $78.48$ & $72.79$	\\ 
\hline
DeepPyramid	\cite{RRanjan_DeepPyramidFD}			& $66.17$ 	& $42.35$ & $71.19$&  $66.07$	\\ 
\hline
HyperFace \cite{RRanjan_Hyperface}				& $90.52$ 	& $90.32$ &  $73.01$& $71.14$	\\ 
\hline
FSFD $C_{\text{best}}$ \cite{FSFD_Mahbub}	& $59.06$	& $55.65$ & $55.74$ & $26.88$	\\ 
\hline
SegFace \cite{SegFaceDeepSegFace_FG2017}					& $67.12$	& $63.09$ & $66.44$ & $61.47$	\\ 
\hline
DeepSegFace \cite{SegFaceDeepSegFace_FG2017}			& $87.16$	& $86.49$ & $82.26$ &	$76.28$ \\ 
\hline
DRUID 			& $\mathbf{91.65}$	& $\mathbf{91.52}$ & $\mathbf{88.59}$ &	$\mathbf{86.88}$ \\ 
\hline
\end{tabular}
\label{Detection_Results}
\vskip -5pt
\end{table}

\subsubsection{Performance}
In table \ref{Detection_Results}, the performance of DRUID is compared with other facial segment-based and state-of-the-arts methods for both datasets in terms of TAR at $1\%$ FAR and recall at $99\%$ precision. From the measures on the AA-01-FD and UMDAA-02-FD datasets, it can be seen that DRUID outperforms all the other detectors on both metrics. In fact, on UMDAA-02-FD datset, DRUID improves the TAR by almost $6\%$ and recall by $10\%$ compared to the second best DeepSegFace. Among the proposal-based methods, SegFace, in spite of being a traditional feature based algorithm, outperforms DCNN-based algorithms such as NPD and DeepPyramid on the AA-01-FD dataset.

\begin{figure}[t]
\centering
\includegraphics[width=0.48\textwidth]{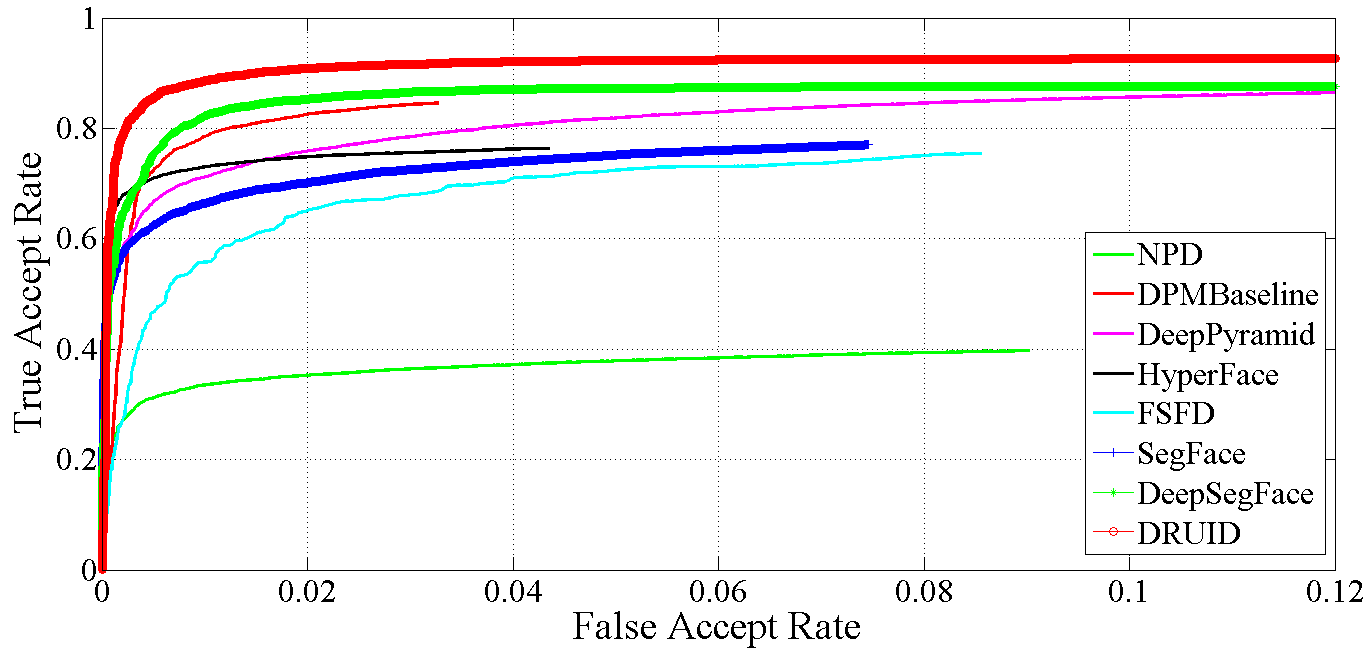}
\caption{ROC curve for comparison of different face detection methods on the UMDAA-02-FD dataset}
\label{ROCUMDAA02}
\end{figure}

\begin{figure}
\centering
\includegraphics[width=0.48\textwidth]{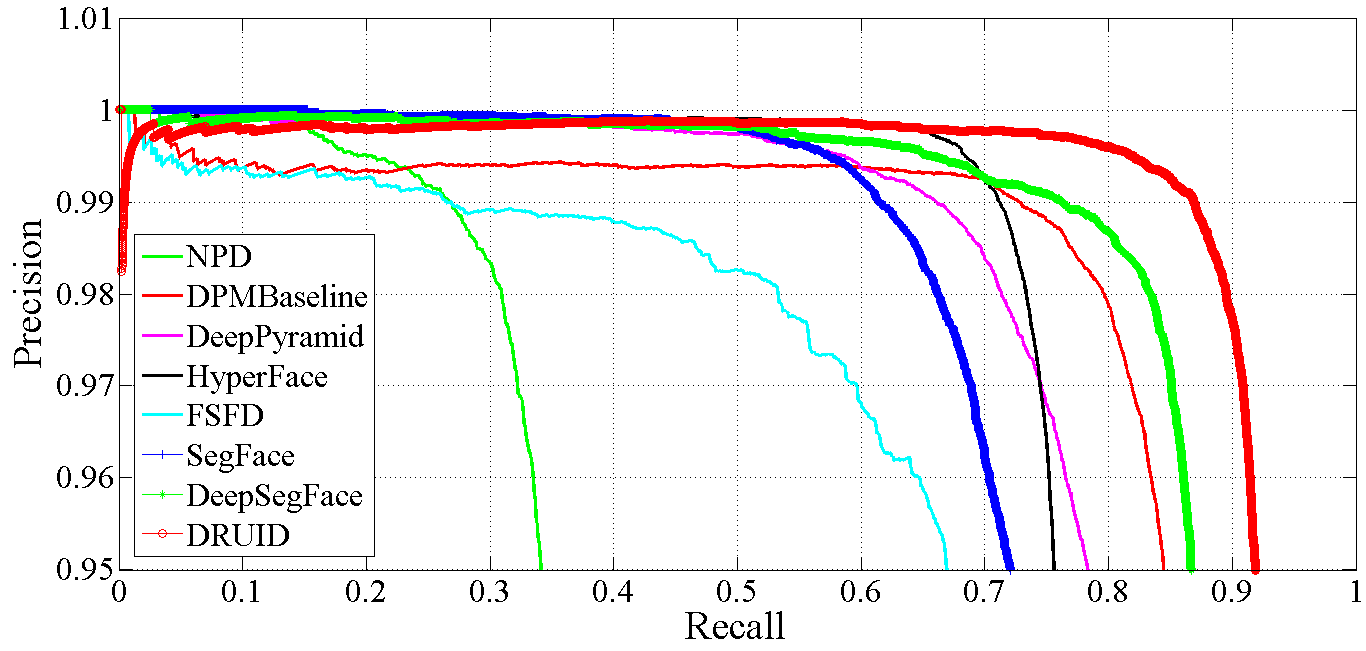}
\caption{Precision-Recall curve for comparison of different face detection methods on the UMDAA-02-FD dataset.}
\label{PrecisionRecallUMDAA02}
\end{figure}

\begin{figure}[t]
\centering
\includegraphics[width=0.48\textwidth]{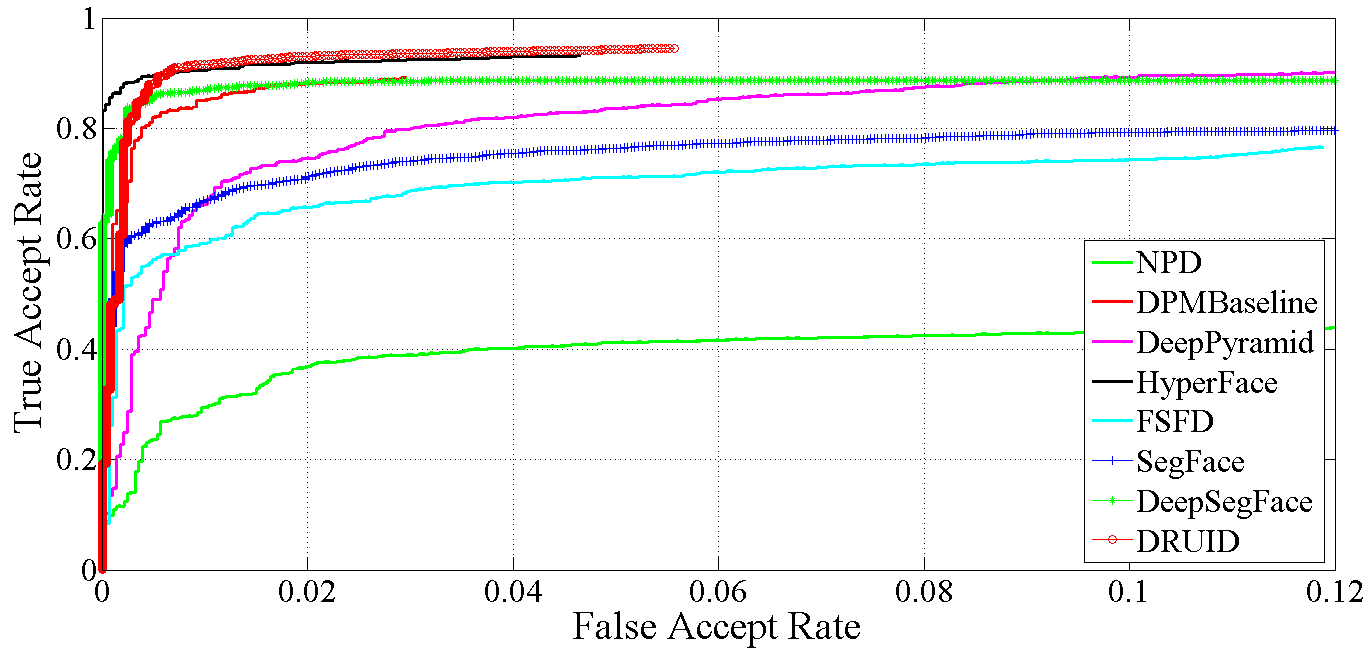}
\caption{ROC curve for comparison of different face detection methods on the AA-01-FD dataset}
\label{ROCUMDAA01}
\end{figure}

\begin{figure}
\centering
\includegraphics[width=0.48\textwidth]{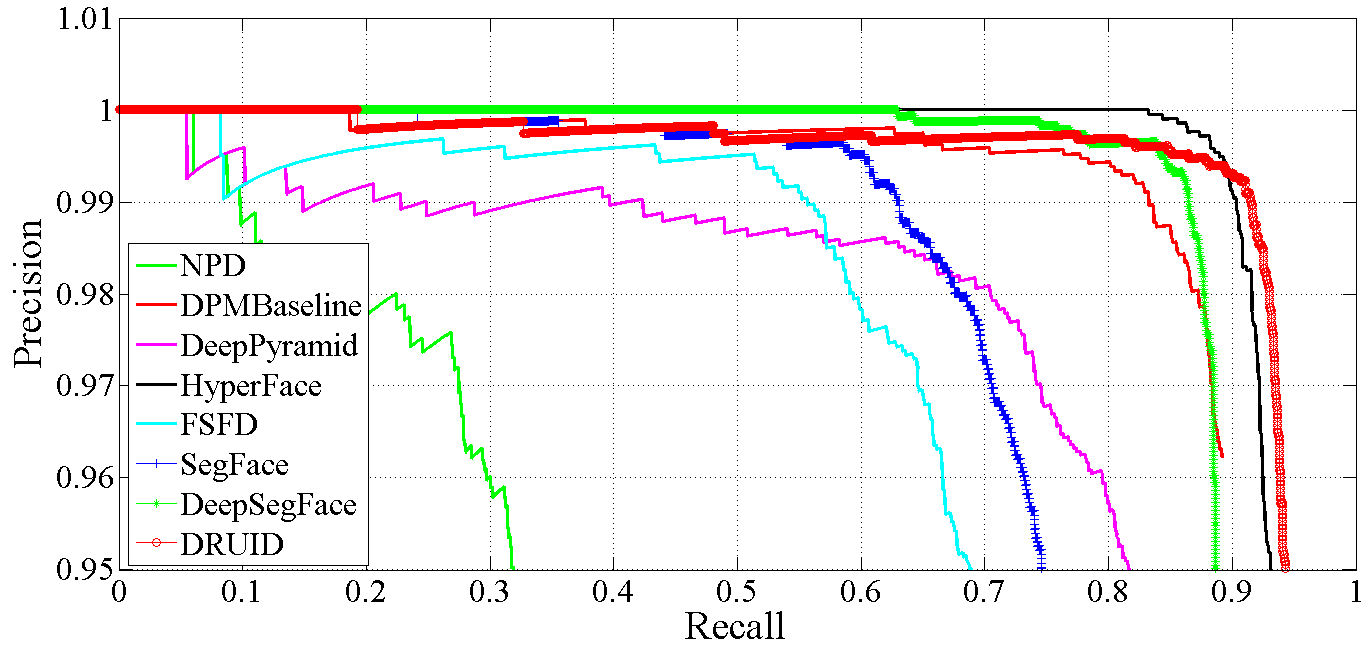}
\caption{Precision-Recall curve for comparison of different face detection methods on the AA-01-FD dataset.}
\label{PrecisionRecallUMDAA01}
\end{figure}

The TAR-FAR and precision-recall curve for UMDAA-02-FD dataset is shown in Figures \ref{PrecisionRecallUMDAA02} and \ref{ROCUMDAA02}. The TAR-FAR and Precision-Recall curves for the AA-01-FD dataset are shown in Figures \ref{ROCUMDAA01} and \ref{PrecisionRecallUMDAA01}, respectively. DRUID outperforms all the other methods in terms of TAR at $1\%$ FAR and recall at $99\%$ precision.

From fig. \ref{ROCUMDAA02}, the ROC for UMDAA-02-FD dataset, it can be seen that DeepSegFace gives the second best performance even with the bottleneck of proposal generation (the curve flattens around $87.5\%$). This is because all the traditional methods perform poorly when detecting mobile faces in truly unconstrained settings that a true acceptance rate of even $87\%$ is hard to achieve. The DRUID network, being free from proposals, achieves $87.45\%$ TAR at $1\%$ FAR.

\begin{figure}
\centering
\includegraphics[width=0.48\textwidth]{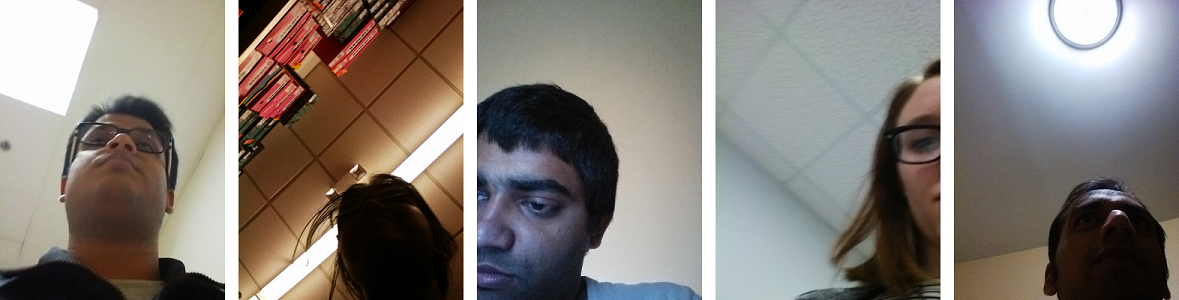}
\caption{Images without even one good proposal returned by the proposal generation mechanism. This bottleneck can be removed by using better proposal generation schemes.}
\label{imagesWithNotEvenOneGoodProposal}
\end{figure}

\begin{figure*}[t]
\centering
\includegraphics[width=0.9\textwidth]{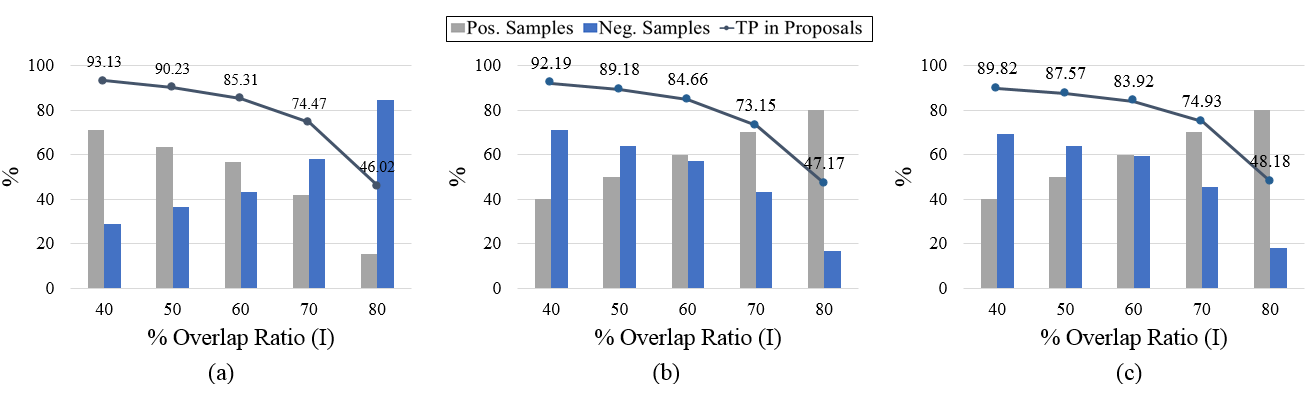}
\vskip -8pt
\caption{(a) for $57756$ Train Proposals from AA-01-FD Dataset, (b) $39168$ Test Proposals  from AA-01-FD Dataset, and (c) $410138$ Test Proposals from UMDAA-02-FD dataset. In all cases $c=2$ and $\zeta=10$.}
\label{InfoBarGraphs}
\end{figure*}

\subsubsection{Performance bottleneck in proposal based techniques}
In Fig. \ref{imagesWithNotEvenOneGoodProposal}, some images are shown for which the proposal generator did not return any proposals or returned proposals without sufficient overlap, even though there are somewhat good, visible faces or facial segments in them. The percentage of true faces represented by at least one proposal in the list of proposals for the training and test sets are counted. The result of this analysis is shown in Fig. \ref{InfoBarGraphs}. The bar graphs denote the percentage of positive samples and negative samples present in the proposal list generated for a certain overlap ratio. For example, out of the $55,756$ proposals generated for training, there are approximately $62\%$ positive samples and $35\%$ negative samples at an overlap ratio of $50\%$. Considering the overlap ratio fixed to $50\%$ for this experiment, it can be seen from the line plot in Fig. \ref{InfoBarGraphs}(b), corresponding to the AA-01-FD test set, that the proposal generator actually represent $89.18\%$ of the true faces successfully and fails to generate a single good proposal for the rest of the images. Hence, the performance of the proposed detectors is upper-bounded by this number on this dataset, a constraint that can be mitigated by using advanced proposal schemes like selective search which generates around $2000$ proposals per image for Hyperface, compared to just around sixteen proposals that are generated by the fast proposal generator employed by DeepSegFace. Another approach to mitigate the bottleneck imposed by the proposal generation step is by modeling the task as a regression problem which is done in the DRUID network. When considering the UMDAA-02-FD test set, which is completely unconstrained and has almost ten times more images than AA-01-FD test set, this upper bound might not be so bad. From Fig. \ref{InfoBarGraphs}(c) it can be seen that the upper bound for UMDAA-02-FD is $87.57\%$ true positive value for the proposal generator and from Table \ref{Detection_Results}, it can be seen that DeepSegFace was able to achieve $82.26\%$ TAR. However, this proves the argument about proposal generation being the bottleneck for this sort of methods, while DRUID, being free from such constraints, is able to demonstrate much better performance in terms of both evaluation measures.

\subsubsection{Performance of auxiliary networks in DRUID}
To investigate the training of the auxiliary networks that predict different facial segments, the TAR-FAR and Precision-Recall curves for each facial segment are plotted in Figs. \ref{TAR_FAR_ROCforAFLWSegments} and \ref{Precision_Recall_ROCforAFLWSegments}, respectively, for the AFLW dataset. Interestingly, it can be seen from these figures that for DRUID the ranking of segments is not similar to the proposal-based methods. The plots reveal that facial segments that represents a bigger portion of the face are trained better. In fact, the nose segment $NS$ which represents the smallest facial segment, performs the poorest, while all the big segments such as $L_{34}$, $R_{34}$, $B_{34}$ etc. are much better trained.

\begin{figure}[t]
\centering
\includegraphics[width=0.48\textwidth]{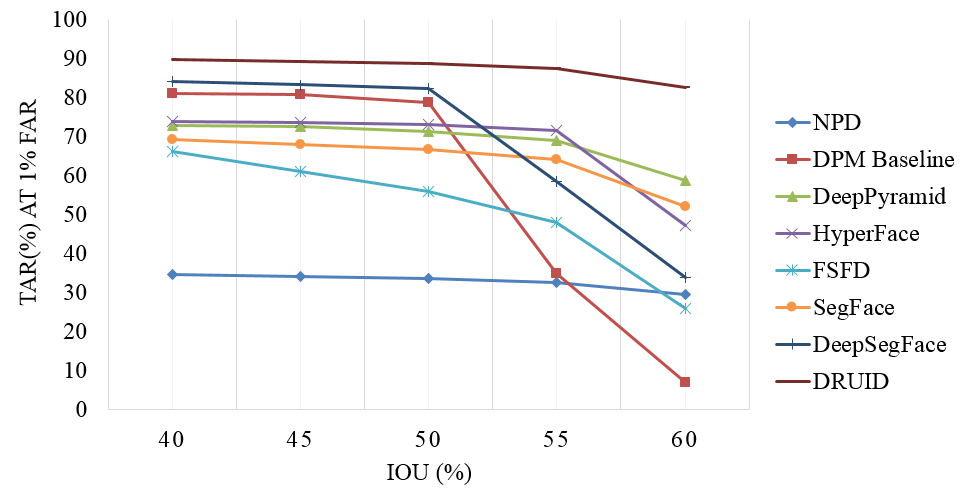}
\caption{TAR at $1\%$ FAR for different IOU for different methods.}
\label{TARAt1PrcntFAR_vs_IOU}
\end{figure}

\begin{figure}[t]
\centering
\includegraphics[width=0.48\textwidth]{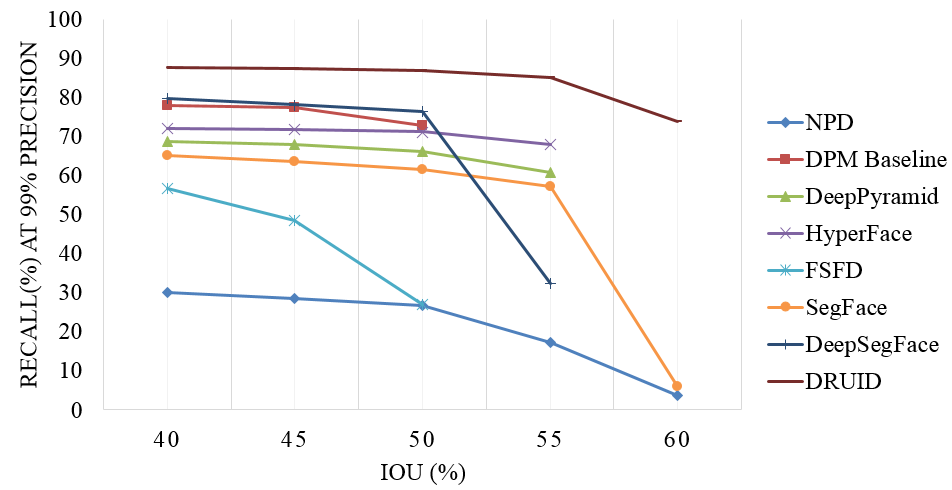}
\caption{Recall at $99\%$ Precision for different IOU for different methods.}
\label{RecallAt99PrcntPrec_vs_IOU}
\end{figure}

\subsubsection{Timing information}
The processing time per image for FSFD on an Intel Xeon(R) CPU E5506 operating at $2.13$GHz with $5.8$GB memory is $0.52$ seconds without any multi-threading or special optimization. SegFace takes around $0.49$ seconds when running on an Intel Xeon CPU E-2623 v4 ($2.604$ GHz) machine with $32$GB Memory without multi-threading, hence it is possible to optimize it to run on mobile devices in reasonable time without requiring a specialized hardware. When forwarding proposals in batch sizes of $256$, DeepSegFace takes around $0.02$ seconds per proposal on a GTX Titan-X GPU. The end to end throughput for DRUID on a Titan-X GPU is also $0.02$ seconds (batch size $32$), which is faster that DeepSegFace since it does not require time for proposal generation. 

For DeepSegFace, the proposals are analyzed to reveal that on an average only three segments per proposal are present for both datasets. Thus, while there are nine convolutional networks in the architecture, only three of them need to fire on an average for generating scores from the proposals.

\subsubsection{Sample detections}
\begin{figure*}
\centering
\includegraphics[width=0.9\textwidth]{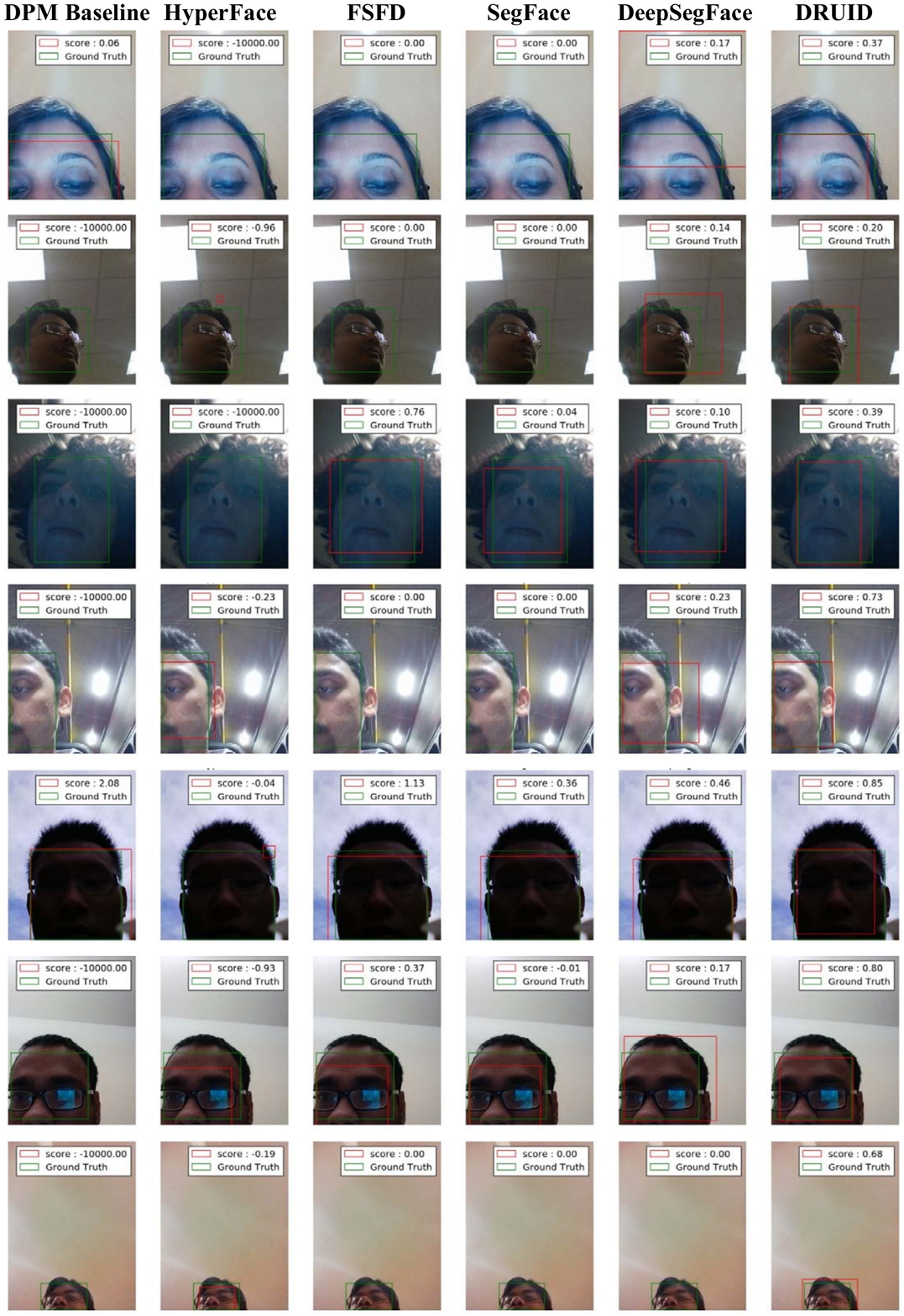}
\caption{Face detection performance comparison of the four facial segment-based detectors with DPM Baseline and HyperFace.}
\label{DetectionComparison1}
\end{figure*}

\begin{figure*}[t]
\centering
\includegraphics[width=0.9\textwidth]{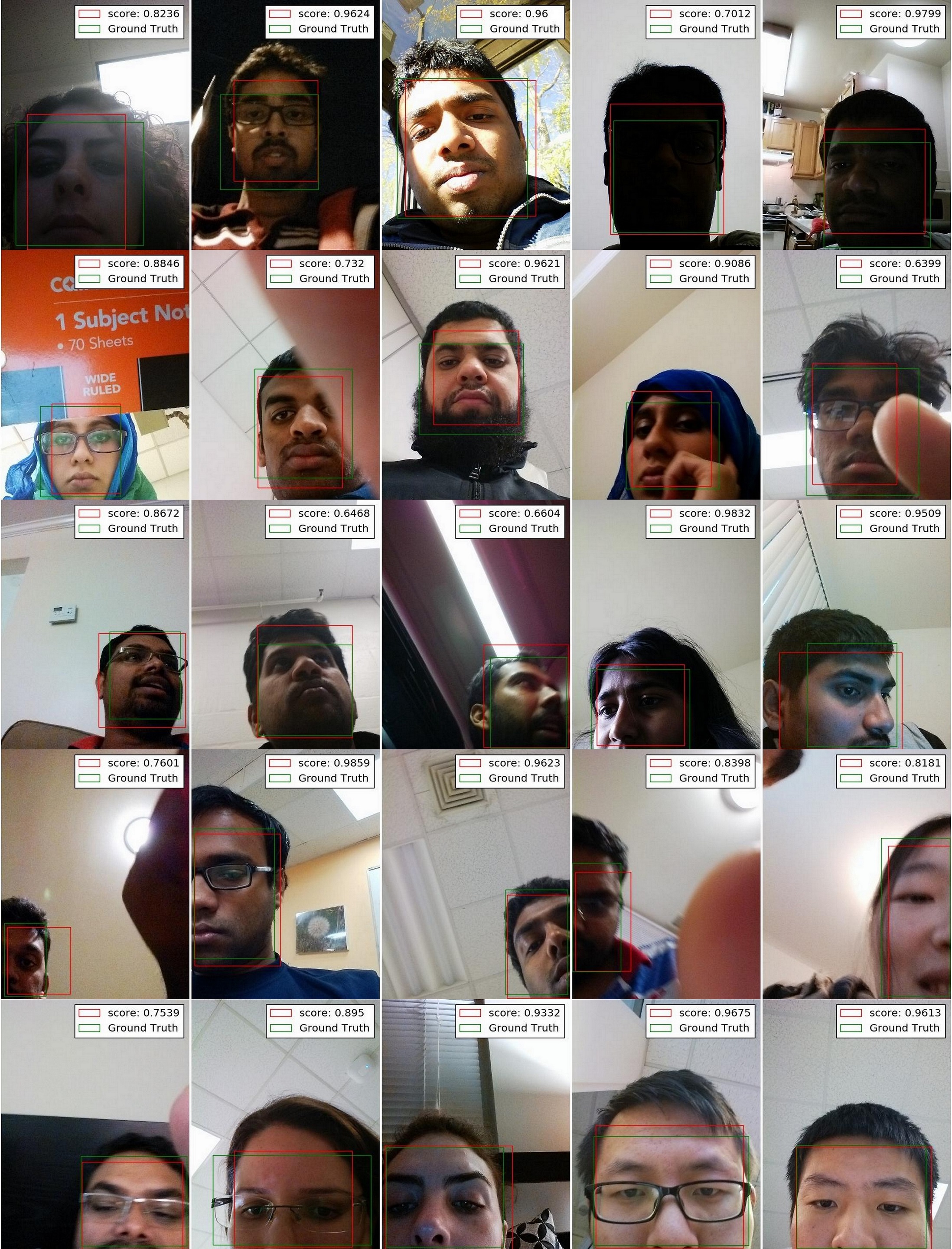}
\caption{Sample face detection outcome of the proposed DRUID method on the UMDAA-02-FD dataset. The five rows (top to bottom) show detection at different illumination, occlusion. pose variation, partial visible side faces and partially visible upper portion of faces, respectively.}
\vskip -8pt
\label{umdaa02_crop}
\end{figure*}

\begin{figure*}[t]
\centering
\includegraphics[width=0.6\textwidth]{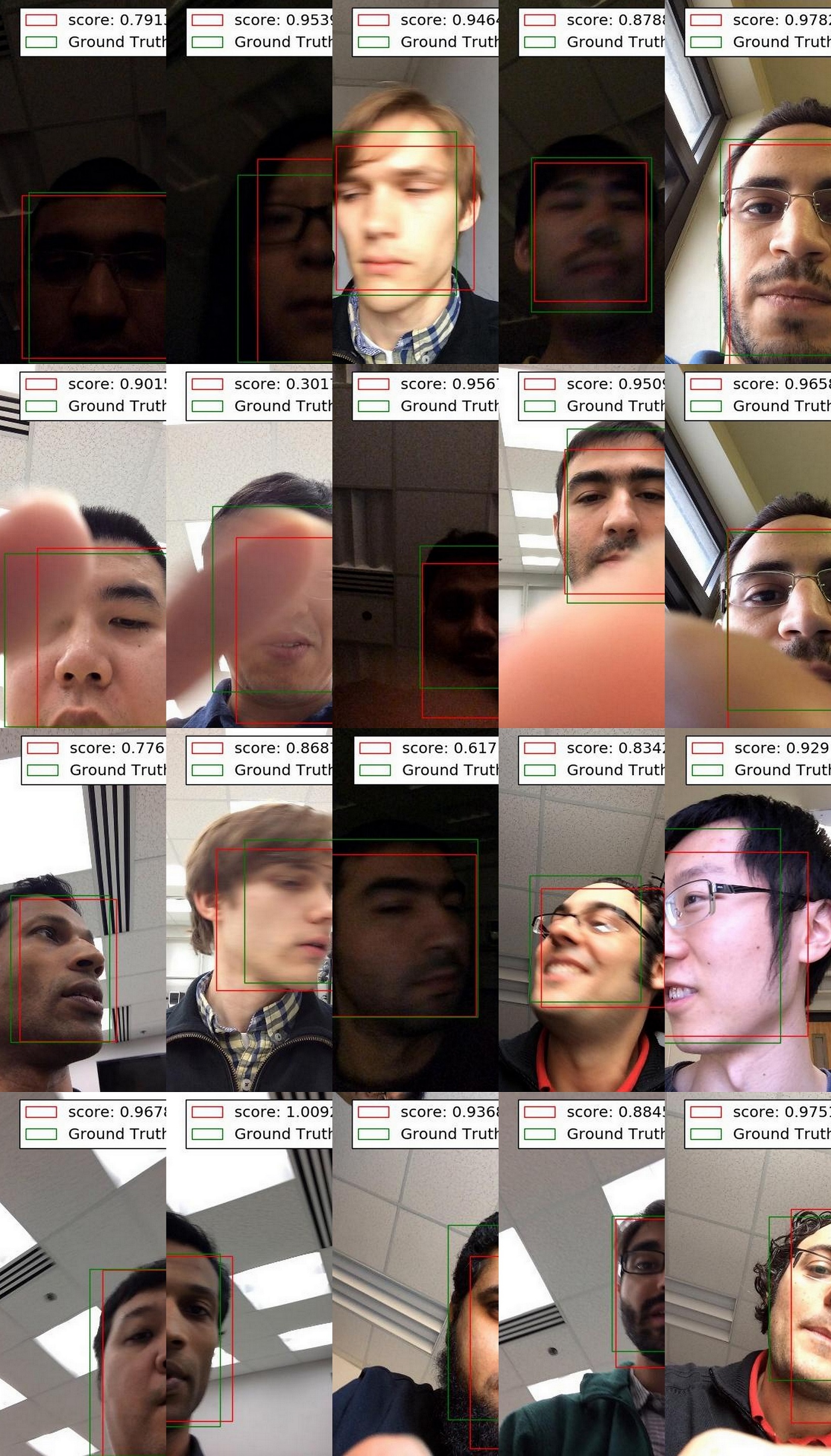}
\caption{Sample face detection outcome of the proposed DRUID method on the UMDAA-02-FD dataset. The four rows (top to bottom) show detection performance at various illumination, occlusion, pose and partial visibility. }
\vskip -8pt
\label{umdaa01_crop}
\end{figure*}

\begin{figure*}[t]
\centering
\includegraphics[width=0.7\textwidth]{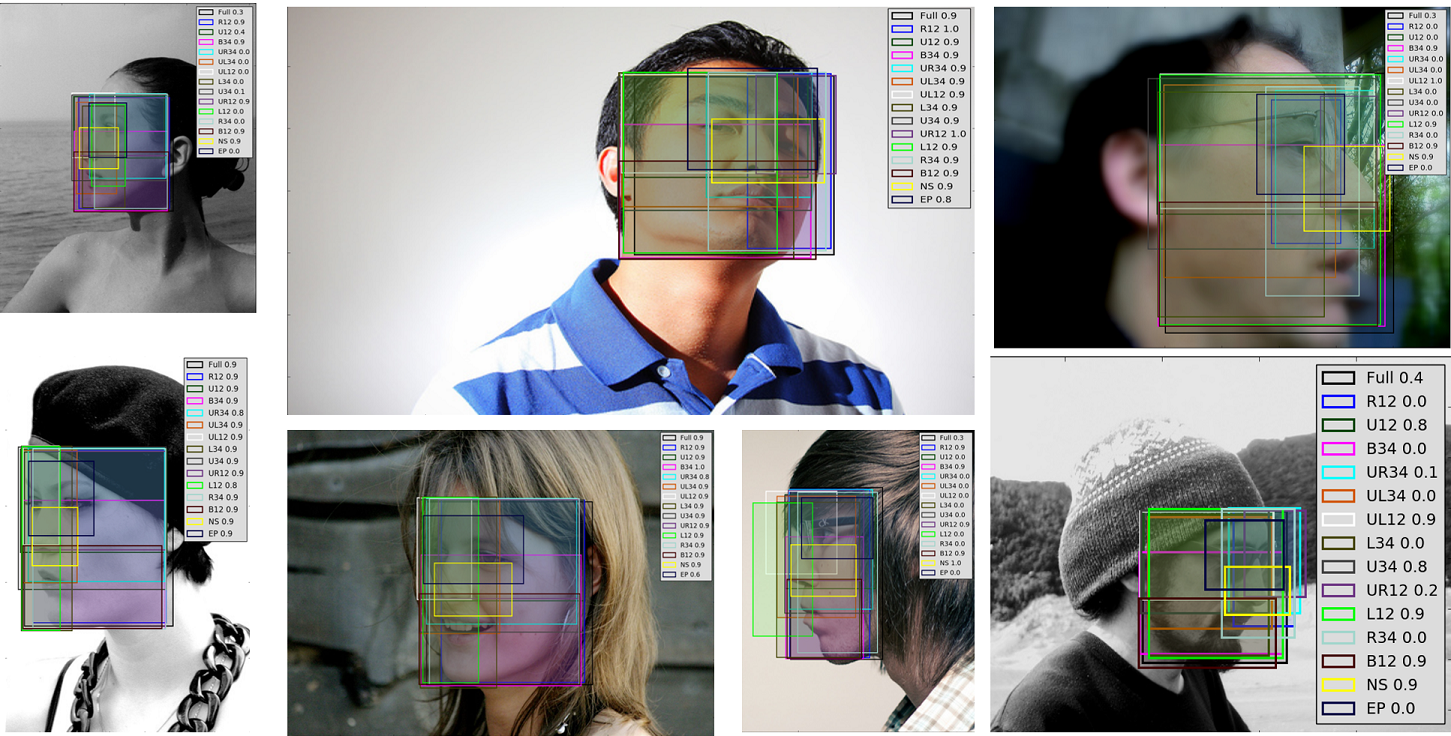}
\caption{Some images from AFLW showing face detections from DRUID. The facial segments bounding boxes and their scores are shown too}
\label{DRUID_faceparts}
\end{figure*}

Some sample face detection results of the DRUID network along with DPM Baseline, HyperFace, FSFD, SegFace and DeepSegFace are shown in Fig. \ref{DetectionComparison1}. It can be seen that, the proposed network performs much better than the others for extreme illumination and pose variation and partial visibility of faces. More sample images depicting the detection performance of DRUID at various illumination, pose and occlusion are provided in Figs. \ref{umdaa02_crop} and \ref{umdaa01_crop} for the UMDAA-02-FD and AA-01-FD datasets, respectively. Fig. \ref{DRUID_faceparts} shows the bounding boxes of facial segments in some images from AFLW. DRUID returned correct bounding boxes with high visibility scores located spatially in appropriate positions. It can be seen that if a face is in profile, then parts that are not visible have very low scores. For example in Fig. \ref{DRUID_faceparts} the top left face's left half is not visible hence left segments like $L_{12}$, $UL_{34}$ and $UL_{12}$ have zero scores.

\begin{figure*}[t]
\centering
\includegraphics[width=0.7\textwidth]{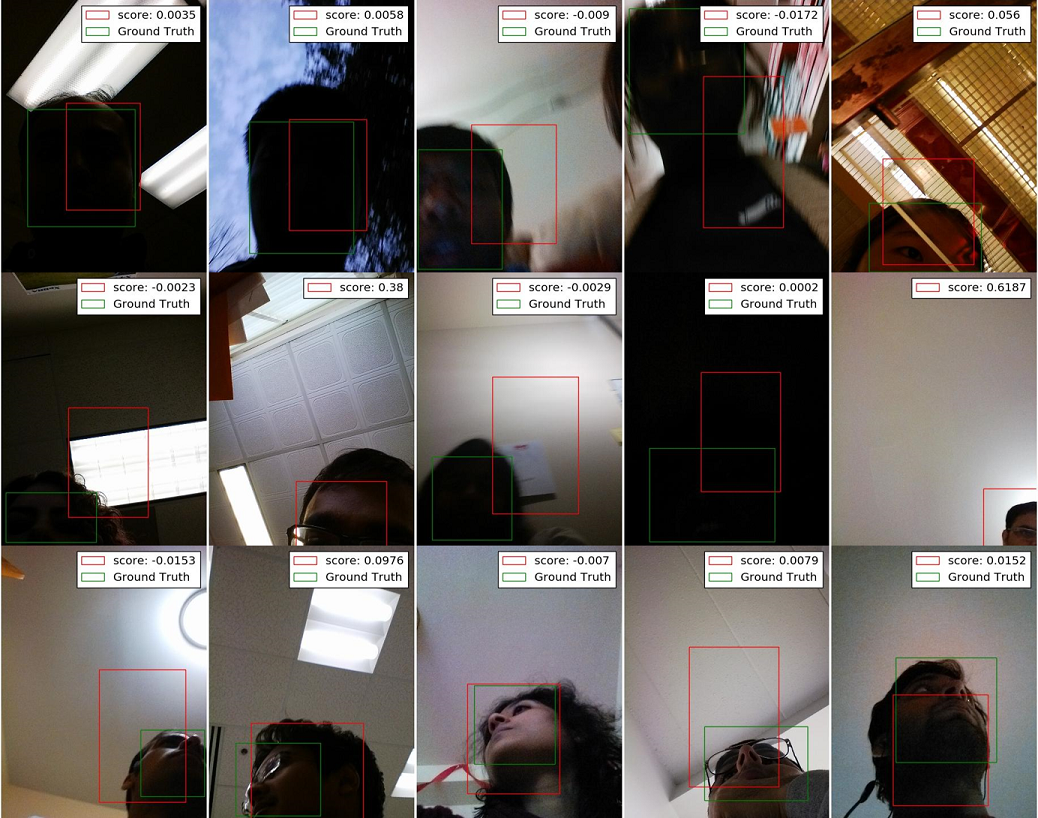}
\caption{Some failure cases of the DRUID network for the UMDAA-02-FD dataset. The first row fails as the images are very blurry, low contrast and difficult in general. The second row shows failures due to very stringent ground-truth annotations that mark near invisible faces or due to cases when the detector detects a partial face that is unannotated in the groundtruth. The third row shows failures due to extreme poses}
\vskip -8pt
\label{bad02}
\end{figure*}

\begin{figure*}[t]
\centering
\includegraphics[width=0.7\textwidth]{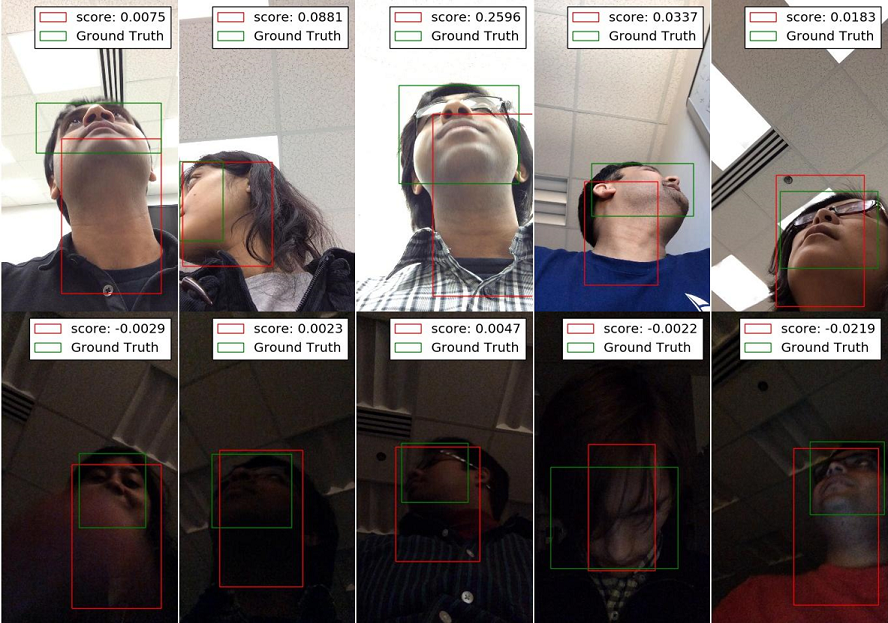}
\caption{Some failure cases of the DRUID network for the AA-01-FD dataset. Most failures can be attributed to extreme poses}
\vskip -8pt
\label{bad01}
\end{figure*}

Some failure cases are presented in Figs. \ref{bad02} and \ref{bad01}. It can be seen from these figures, that the network failed mostly in extremely difficult scenarios. In some cases, the ground truth (green) is questionable, for example, in Fig. \ref{bad02}, image $2$ and $4$ of row $2$. The first one does not have a proper ground truth while in the second one, the face is not visible at all but a ground truth is marked. 

\section{Conclusion and Future Directions}\label{Conclusion}
This paper proposes DRUID, a deep regression-based scheme for efficiently handling the problem of detecting partially visible faces which are prevalent in the mobile face domain. The proposed method is compared with three other facial segment-based face detection methods, namely, FSFD, SegFace and DeepSegFace, developed previously, which rely on a proposal generator unlike DRUID. In this regard, the principal differences between proposal-based and regression-based face detection are elaborated here and also the improved performance of these detectors over state-of-the-art face detectors in the mobile domain is demonstrated on two different mobile face dataset. The paper also covers the novel data augmentation and regularization techniques adopted for DeepSegFace and DRUID as well as the unique drop-out mechanism achieved by using facial segment-based proposals. The DRUID network is found to outperform all the other methods by a wide margin on the UMDAA-02-FD and AA-01-FD datasets.

The idea of regression-based single face and facial segment detection can be extended to multiple face and facial segments detection following the principle of \cite{liu2016ssd}. The detection of facial segments opens up the research opportunity to investigate facial segment-based age, gender, pose and other attribute detection and face verification. Future research  direction may also include customizing and optimizing these algorithms to make them suitable for real-time application on mobile-platforms.

\section*{Acknowledgement}
This work was done in partnership with and supported by Google Advanced Technology and Projects (ATAP), a Skunk Works-inspired team chartered to deliver breakthrough innovations with end-to-end product development based on cutting edge research and a cooperative agreement FA8750-13-2-0279 from DARPA.

{\small
\bibliographystyle{IEEEtran}
\bibliography{biblio_PFD}
}

\end{document}